\documentclass[fleqn,10pt]{wlscirep}
\usepackage{hyphenat}
\usepackage[utf8]{inputenc}
\usepackage[T1]{fontenc}
\usepackage{tabularx}
\usepackage{dsfont}
\usepackage{bbm}
\usepackage[toc]{appendix}
\usepackage{dsfont}
\usepackage{hyperref}
\usepackage{amsmath}
\usepackage{microtype}

\usepackage{algorithm}
\usepackage{algpseudocode}
\usepackage{amsmath}
\title{Columnar-Embedder: A Biologically Inspired Cortical Architecture for Binary Sparse Distributed Graph Representations}

\author[1,*]{Mohamed Abidalrekab}
\author[2]{Dan Hammerstrom}

\affil[1]{Portland State University, Electrical and Computer Engineering, Portland, 97223, USA}
\affil[2]{Portland State University, Electrical and Computer Engineering, Portland, 97223, USA}

\affil[*]{Moh29@pdx.edu, dwh@pdx.edu}

\begin{abstract}
Finding a representative description of graph entities that captures their structural roles and homophily is a challenging goal for graph embedding techniques due to the non-Euclidean nature of graphs. Traditionally, Graph embeddings achieve top performance via random-walk methods and graph neural networks. However, these methods are transductive and utilize an expensive global optimization via softmax or a dense representation trained in an end-to-end pipeline with gradient descent. Nonetheless, other variants of GNNs can map to unseen nodes; they still rely on iterative message passing and backpropagation, incurring high computational and memory costs. Conversely, the mammalian cortex solves structurally similar problems by learning to map its input stream of patterns into a compact representation for downstream regions. We present the biologically inspired Columnar-Embedder architecture for learning binary Sparse Distributed Representations (SDRs) of graph nodes. The learning is driven by a local Bienenstock-Cooper-Munro (BCM) Hebbian rule modulated by positive pointwise mutual information (PPMI) computed from online streams of random walks. Continuous learning from streaming random-walk pairs without labels, backpropagation, or supervision enables the architecture to exhibit natural resistance to catastrophic forgetting. Across five graph benchmarks, the performance of SDRs is competitive with that of real-valued dense embeddings on node classification and link prediction, while the architecture exhibits portability, resilience to noise, and robustness to data corruption. 
\end{abstract}

\begin{document}

\flushbottom

\keywords{Cortical columnar, Modulated-BCM, sparse coding, Homeostasis, winner-take-all, graph embedding, Intrinsic plasticity, Outstar rule, Anti-Cross Entropy, SDR}
\maketitle
\thispagestyle{empty}

\section*{Introduction}

Graphs are a natural representation tool for relational data sets such as citation networks, social networks, molecular drug \& protein structures, knowledge bases, recommendation systems, and many others. However, the non-Euclidean nature of graphs makes it difficult to extract useful representations or to apply learning algorithms directly. Thus, learning to embed graph nodes and edges into a high-dimensional vector space enables the graph to be applied to many downstream tasks \cite{Zhou2018GraphNN, GraphEmbeddingReview}. The downstream applications span node classification \cite{Bhagat2011NodeCI}, link prediction \cite{LinkPredictionBasedonGNN}, graph clustering \cite{Graphclusterng}, temporal recommendations \cite{Shen2024TemporalGN}, question answering over knowledge graphs \cite{Yasunaga2021QAGNNRW}, drug-target affinity \cite{Voitsitskyi20233DProtDTAAD}, molecular property prediction \cite{Liu2018ChemiNetAM}, and financial market modeling \cite{wu2025trading} among many others. Although applications appear to span a wide range, the methods that produce these embeddings exhibit striking architectural homogeneity.

\par Initially, researchers developed graph embedding algorithms \cite{GraphEmbeddingReview} that relied on direct knowledge of graph properties to handcraft the features for each node. This requires deeper analysis of graph structural components, using an adjacency matrix and other related information. However, these techniques are deemed impractical for two reasons: the high computational and memory costs when applied to large-scale graphs, and poor generalization due to construction requirements that are specific to particular graphs. Recently, a broad taxonomy of graph embedding methods discussed in the literature points to three major categories: (i) Matrix-factorization based methods, (ii) Random walks based methods, and (iii) Neural Network based methods. However, this work introduces a fourth category of Bio-inspired architectures that employ a sparse binary representation. 

\par Matrix factorization-based methods (e.g., \cite{belkin2001laplacian,roweis2000nonlinear, cao2015grarep}) construct a high-order proximity matrix from transition probabilities and factorize it to obtain node embedding. Still, they do not scale well to large networks and rely on computationally expensive matrix decomposition. Their key innovation is the use of an adjacency matrix to infer node similarity and relationships from the presence of an edge. For instance, Locally Linear Embedding \cite{roweis2000nonlinear}, Laplacian Eigenmaps \cite{belkin2001laplacian}, HOPE \cite{HOPE}, and the Stochastic Neighborhood Embedding (SNE) method \cite{SNE} derive their similarity measures from factorizing the adjacency matrix, which led to good performance in the graph reconstruction framework. However, those methods are inherently prone to overfitting the graph's original adjacency matrix because they rely solely on immediate local connections, thereby degrading embedding quality. Thus, the output of the embedding step may not be suitable for specific downstream tasks, such as link prediction. Furthermore, due to their computational complexity of $O(|V|^2)$, adjacency matrix-based measures are difficult to scale to large networks.

\par In this paper, we focus on scalable models that inherit key features from random-walk and neural-network methods. The groundwork for graph embedding was laid by the Skip-Gram model, which was introduced by Mikolov et al. \cite{Mikolov2013DistributedRO} in the field of natural language processing. The Skip-Gram model is an end-to-end framework that learns high-quality distributed vector representations of words via a specific context window. It captures numerous precise syntactic and semantic relationships among words in a text corpus. Distributed representations of words in a vector space help learning algorithms achieve better performance in natural language processing tasks by grouping similar words within a similar context window \cite{Mikolov2013EfficientEO}. Random walk techniques began with DeepWalk \cite{DeepWalk} and were extended by Node2vec (n2v) \cite{Grover2016node2vecSF} and LINE \cite{LINE}, which reformulate the embedding problem as a language-modeling task. Therefore, random walk sequences represent sentences, and the Skip-Gram model \cite{Mikolov2013DistributedRO} is applied to learn the node embeddings. Subsequently, Qiu et al.\cite{Qiu2017NetworkEA} showed that these random-walk methods implicitly factorize functions of the graph transition matrix, placing them in a low-rank family of spectral methods. Although n2v \cite{Grover2016node2vecSF} is a superior method with flexibility and customization, high performance, Scalability, and Parallelization, it suffers from an inability to generalize to New Nodes (Inductive Learning). We adopted a similar pipeline because the biased random walk used by n2v enables graph traversal without applying any expensive operators or transformations. We then mitigate the inductive learning issue by adopting an encoder that mimics the graph neural network (GNN) framework \cite{Scarselli2009TheGN}.

\par Graph neural networks tackle the issue differently: rather than factorizing a similarity matrix or depending on random walks, they learn an embedding by aggregating representative features from each node's local neighborhood. The Graph Neural Network (GNN) was introduced by Scarselli et al. \cite{Scarselli2009TheGN}, extending existing neural networks to process data represented in graph domains. Unlike the n2v method, which does not require nodes to have features to form a representation of a node, GNNs depend on incorporating node features and their neighborhood features to be aggregated and fused to generate a new representation for every node in a graph. There are many variants of GNNs: spectral GCN \cite{GCN}, GraphSAGE \cite{Hamilton2017InductiveRL}, attention-based GAT \cite{Velickovic2017GraphAN, Awedat2025SuperiorGATGA}, and the message-passing framework of Gilmer et al. \cite{Gilmer2017NeuralMP}. Despite the diverse training objectives and aggregation schemes, these methods share 3 drawback properties: (1) dense real-valued high-dimensional representations of size 64-512, (2) globally coordinated gradient descent training via softmax over node vocabulary or backpropagation through message-passing layers, and (3) a batched, in-memory view of the graph during the training. In addition, their performance degrades due to scalability issues, limited structural expressiveness, and information loss. Recent work on binarized neural networks \cite{Courbariaux2016BinarizedNN, Lin2017TowardsAB} and graph-specific Binary hashing \cite{Li2011HashingAF} has shown that the dense-vector assumption is not strictly necessary and that Binary can achieve comparable performance. More strikingly, sparse binary state-space (spiking) models have been shown to match or exceed the transformer architecture \cite{Vaswani2017AttentionIA} on long-range sequence benchmarks \cite{Stan2024Learning}. Yet, they retain global gradient-based training and require a surrogate gradient to handle the non-differentiable binarization step.

\par The mammalian cortex provides valuable insight into how a modular, multilayer architecture maps input stimuli to a compact representation useful for downstream sub-regions. Primary sensory cortices contain feature-dense layers for rapid feedforward processing of incoming data. Cortical representations are sparse and binary-like, with any one stimulus activating only (~1 to 4\%) of available neurons \cite{OLSHAUSEN2004, Olsen2010DivisiveNI, Beyeler2019NeuralCO}. The mathematical properties of sparse distributed representations (SDRs) enable exponential representational capacity, robust pattern completion in the presence of noise, and near-orthogonal random codes that reduce false matches. SDRs have been formalized by Hawkins \& Ahmad \cite{Ahmad2015PropertiesOS, WHySparse} and form the theoretical foundation of Numenta's Hierarchical Temporal Memory (HTM) framework \cite{applicationsHTM, Dauletkhanuly2020, SptialPooler}. The learning process within the cortical architecture employs the Hebbian associative rule, in which neurons that fire together wire together. However, more sophisticated learning rules have been observed in biology \cite{Bienenstock1982TheoryFT, Izhikevich2003RelatingST, Clopath2010ConnectivityRC} that help to form specific, sparse activity patterns within a local layer and adapt to changes in input statistics. The Bienenstock-Cooper-Munro (\textbf{BCM}) rule \cite{Bienenstock1982TheoryFT, Cooper2012TheBT} is a quadratic Hebbian rule with a sliding metaplastic threshold, which explains how BCM provides a model for how high-frequency stimulation leads to long-term potentiation (LTP) and low-frequency stimulation leads to long-term depression (LTD). A voltage-based STDP \cite{Clopath2009Connectivity} unifies spike time-dependent plasticity (STDP), voltage-dependence, and frequency-dependence in a single biophysical rule. Practically, most local Hebbian learning rules tend to saturate synaptic weights due to excessive firing. Therefore, intrinsic plasticity, which is a form of Homeostasis \cite{Carlson2013BiologicallyPM}, adapts each neuron's excitability to maintain target firing rates over long timescales \cite{Turrigiano2012HomeostaticSP}. Together with biological neuron models such as the Adaptive-Exponential leaky-Integrate-and-Fire neuron model (AdEx-LIF) \cite{Brette2005, Naud2008FiringPI}, and the broader cortical columnar architecture \cite{Lundqvist2006, johansson2005mean, Kaplan2014ASN}, these ingredients support fast few-shot learning, continuous learning without catastrophic forgetting \cite{Iyer2021AvoidingCA}, and noise-resilient pattern completion. Spike-based computation on neuromorphic hardware \cite{Merolla2014AMS, Esser2015BackpropagationFE} offers an order-of-magnitude of energy savings over dense gradient-based deep learning, especially for graph traversal and random walk workloads \cite{Smith2020NeuromorphicSA}. 

\par Existing works on graph embedding that utilize biologically inspired architectures combining spiking neural nets with graphs have largely focused on downstream applications rather than on embedding as a means of constructing node and edge representations. Zhu et al \cite{zhu2022spiking} introduce a more general SpikingGCN that approximates the traditional GCN network \cite{GCN} via surrogate-gradient training. SpikingGCN encodes graph-convolved features into spike trains using Bernoulli coding and reads out per-neuron firing rates to be used on node classification tasks. Dy-SIGN \cite{10.1609/aaai.v38i15.29587}  adds an information-compensation channel to recover details lost due to spiking, and SiGNN uses spikes only to gate continuous features rather than to replace them. Furthermore, SiGNN \cite{Chen2024SiGNNAS} extends spiking GNNs to dynamic graphs, and SSEL \cite{Yang2025SSELSS} combines event-driven sparsity with a structural-entropy objective for adversarial robustness on citation graphs. The TactileSGNet line \cite{Gu2020TactileSGNetAS, Yang2023AMSGCNTO, Yu2024G2TSNNFT, Guo2025EventDrivenTS} applies spiking graph convolution networks to event-based tactile recognition. These are important first steps, yet they lack bio-inspired mechanisms, as they retain the message-passing GCN pipeline and produce dense, continuous, real-valued representations. Additionally, SpikingGCN is still gradient-trained at the weight level via surrogate gradients due to its non-differentiable activation function. It operates in supervised classification settings rather than producing reusable unsupervised node embeddings. The hardware benchmarks MiniSeer and MicroSeer \cite{Zhu2026CitationND} are built on the STDP-based neuromorphic graph-learning approach \cite{Cong2023HyperparameterOA}, which is the closest genuine combination of spiking nets and graphs. The approach uses spike-timing-dependent plasticity with no gradient training at all, yet propagates known labels through a spiking network whose topology mirrors the citation graph, rather than learning a reusable, inductive node embedding. Nevertheless, mirroring the topology scheme raises serious concerns about the scalability to large graphs. To the best of our knowledge, no prior work has applied a strictly biological cortical architecture, sparse columnar connectivity, BCM Hebbian learning, intrinsic plasticity, and a local diversification rule to  graph-embedding benchmarks. 

\par This work asks: Can a sparse, locally trained, biologically inspired cortical architecture produce graph node embeddings whose downstream usefulness is competitive with that of dense, gradient-trained baselines on downstream tasks? The answer we believe is yes. The proposed Columnar-Embedder architecture produces a very compact, competitive, and resilient representation that achieves results matching those of state-of-the-art methods on node classification and link prediction tasks. The Columnar-Embedder architecture embeds graphs by translating random-walk sequence statistics into proportional overlaps in node representations, ensuring that nodes with similar neighborhoods get similar representations. The architecture operates under strict constraints of representational and connectivity sparsity, local continual Hebbian learning, and no dependence on external information other than the graph adjacency matrix. Thus, our work presents five main contributions that distinguish it from previous attempts.
\begin{enumerate}
    \item A three-layer cortical architecture that is inspired by the neocortex and other biological sensory modules.
    \item The BCM-PPMI local Hebbian learning rule modulated by online random-walk co-occurrence statistics. 
    \item The Anti-Cross-Entropy (Anti-CE) plasticity rule for within-column neuron diversification.
    \item A fully portable architecture that scales to larger datasets , with no modification or extra tuning.
    \item Competitive node classification and link prediction results on Cora, Citeseer, and Large Amazon Photo datasets against n2v, DeepWalk, LINE, supervised GCN, and GraphSAGE. The results are achieved without labels, node features, backpropagation learning, or supervision.
\end{enumerate}
The paper is organized as follows. \S 1 introduces the relevant topics from the literature, \S 2 details the Biological inspiration behind some architectural design decisions, and \S 3 presents the results, including major accomplishments in downstream tasks and representation evaluation. \S 4 presents a general discussion of key aspects of this novel architecture, and \S 5 presents a full presentation of the detailed architecture and its related mechanisms.   

\section*{Biological Inspiration of Our Work}

\par Sparse coding for simple neurons of region V1 in the mammalian sensory cortex was first characterized by Olshausen and Field \cite{Olshausen1996}. Such sparse code formation has been observed virtually in every cortical area where any single stimulus activates only a fraction (typically 1-5\%) \cite{OLSHAUSEN2004} of the neuronal population, evoking many other encoding responses. The empirical advantage of the neuronal coding scheme is well documented in the literature, which includes high representational capacity \cite{WHySparse}, robustness to noise and dropout \cite{Grieves2017TheRO}, energy efficiency due to a low average firing rate of ~5 - 8 Hz, and addressable retrieval through partial-pattern completion \cite{WHySparse, Ahmad2016HowDN, Beyeler2019NeuralCO}. The discrete, event-based, all-or-none nature of spike-based computations makes binary sparse representation suitable for hierarchical processing in cortical circuitry. 
\par Famously, the HTM framework was introduced by Numenta \cite{applicationsHTM} to formalize these properties within an algorithmic architecture comprising cortical mini-columns and an encoding stage. The implementation of local Hebbian learning rules with plasticity-based permanence allows the cortical architecture to behave as a sparse distributed attractor memory \cite{BCPNN}. Hawkins et al. \cite{SptialPooler} use the HTM spatial Pooler (SP) to construct useful SDRs for processing by a subsequent stage of an associative memory to predict temporal sequences, anomalies, or higher-level classifications. The SP mimics neocortex-like online sparse distributed coding that maintains binary SDRs through a biological voting-majority competitive mechanism called K-winner-take-all (k-WTA) across a region of mini-columns \cite{WTA_computation_proof}. Our work inherits both columnar organization and the k-WTA mechanism to perform a hard selection of hyperactive neuronal populations for specific inputs. The formation of Winner-Take-All is analogous to the attention mechanism in Deep learning; However, it develops through local learning without a reference signal or supervision. 

\par The Bienenstock-Cooper-Munro (BCM) rule \cite{Bienenstock1982TheoryFT} is the  Hebbian learning rule, characterized by the expression of the time-evolving synaptic weight as a Hebbian-like product of presynaptic activity and a nonlinear function of postsynaptic activity. The learning rule produces long-term potentiation (LTP) if postsynaptic activity exceeds a sliding metaplastic threshold that represents the time-averaged firing rates and long-term depression (LTD) when activity falls below it. \cite{Intrator1992InvitedAO} extends the original (BCM) model to show that an unsupervised biological neuron can perform complex, high-dimensional feature extraction by finding projections that emphasize non-Gaussian distributions (such as multi-modality). Generally, the BCM learning rule has been shown to develop selectivity for natural stimulus statistics in V1 and other sensory areas and to provide a stable Hebbian dynamic that does not require explicit weight bounds, as it adaptively modulates activity via a sliding window. However, to ensure the stability of neural activity and prevent weight drifting and saturation, a homeostatic regulatory mechanism, such as intrinsic plasticity, IP \cite{Desai1999PlasticityIT}, coupled with (BCM), targets firing rates in the presence of changing input statistics \cite{Cai2024BCMInspiredSC, Cooper2012TheBT}. Intrinsic plasticity is a consistent change in a neuron’s intrinsic electrical properties induced by neuronal or synaptic activity, rather than an alteration in Synaptic efficacy. The three-factor Hebbian learning \cite{Frmaux2016NeuromodulatedSP} enhances BCM with a global modulator term, which can be interpreted as a neuromodulatory signal ( dopamine, noradrenaline, or acetylcholine). The role of neuromodulatory signals in gating synaptic plasticity is based on a behavioral or information-theoretic quantity that spreads across the network. By design, the third factor, or a modulator, does not require a backward pass through the network because it is available at every synapse, either as a scalar or vector. The connection between BCM and Hebbian rules is well known \cite{Izhikevich2003RelatingST} to follow directly from STDP when pre- and postsynaptic neurons have uncorrelated or weakly correlated Poisson spike trains. However, sometimes it is desirable to diversify the representations of a population by weakening the connections between co-active neurons rather than strengthening them, which can be accomplished via Anti-Hebbian rules as introduced by Földiák \cite{Fldik1990FormingSR}. For instance, Columnar-Embedder incorporates the Anti-Cross Entropy (Anti-CE) learning rule within cortical columns to encourage decorrelated populations to form diverse neuronal responses to the same stimulus.

\par Equipped with these locally Hebbian learning rules, no backward passes, no chain rules, no error gradient propagation through layers, no requirement for differentiability, most biologically inspired architectures allow for rich behavioral performance, computational efficiency, and energy savings. The Hopfield \cite{Hopfield1982} network, introduced a recurrent network of binary neurons trained by the local Hebbian outer-product rule, demonstrates the ability of simple local plasticity to produce content-addressable associative memory with attractor dynamics. Furthermore, Willshaw et al. \cite{Willshaw1969NonHolographicAM} extend Hopfield to build a binary associative memory via only sparse binary patterns. These networks serve as the basis for establishing capacity-per-synapse bounds that later motivate sparse coding in much larger architectures \cite{Palm2013NeuralAM}.

\par One of the main contributors to our work is the Bayesian Confidence Propagation Neural Network (BCPNN), a family of models developed by Lansner et al \cite{BCPNN, Lundqvist2006, Kaplan2014ASN, johansson2005mean} that is particularly relevant here. First, cortical mini-columns serve as a discrete unit of attractor memory \cite{1000synapese}, and second, learning is driven by the log-probability ratio of pre- and post- synaptic activities, which is exactly the positive pointwise mutual information PPMI signal used as a third factor of BCM. The PPMI is a statistical count of all pairs of nodes that coincide in the streamed random walks. It serves as a prior for BCM learning-rule computations in a Bayesian framework. Yang et al. \cite{Yang2022HeterogeneousES} further demonstrate that cortical ensembles of the BCPNN style in the spiking architecture support spike-driven few-shot online learning as observed in our work. Structurally, the mammalian piriform cortex and the insect mushroom body consist of similar functionality: a sparse, high-dimensional expansion layer that processes projections from a much smaller input space and disentangles the representations through random or learned projections that adapt to changing input statistics \cite{Kaplan2014ASN}. Our work has a similar organizational structure, forming an explanation layer that increases the combinatorial capacity of the cortical columnar, providing ample headroom to disambiguate any residual duplicate codes at the columnar level. Although biological algorithms largely inspire our work and we seek to mimic them, we had to trade off some biological realism for computational speed without compromising performance. We adopted an architecture with reasonable computational cost and rich detail to demonstrate the scalability, portability, and resilience of the proposed architecture. In addition, the training scheme with online continual learning is designed to accommodate future work on dynamic graphs and nodes with features. 
\section*{Results}

\subsection*{Empirical results}

Given that the primary objective of the proposed architecture is to construct an appropriate representation for graph nodes that captures both their structural role and homophilic relationships, we pursued three complementary evaluation strategies to assess its overall performance and the quality of the resulting representations on undirected, featureless, static graphs of varying sizes. 
\par First, we assessed the quality of the Sparse Distributed Representation (SDR) produced by Columnar-Embedder in two node-level and edge-level tasks, namely node classification and link prediction. These experiments demonstrate the ability of the Columnar-Embedder to map node proximity in the original graph to a proportional degree of overlap between SDRs in the representation space. Second, we investigate intrinsic SDR properties related to geometric preservation and community separability under bit-flip perturbations to understand the robustness of the induced representational geometry. Third, we evaluate the resilience of the Columnar-Embedder architecture to noise, missing data, and corruption, which is essential for identifying potential failure modes and structural weaknesses in the model. Fourth, we demonstrate the architecture capability to scale with no structural or hyperparameters tuning for larger graphs with different characteristics. 
\subsubsection*{Datasets and Experimental protocol}
We evaluate the quality of SDRs produced by the Columnar-Embedder architecture and compare them with baseline methods. As shown in Table \ref{tab:Benchmark}, the three main datasets {CiteSeer, Cora, and Amazon-Photo} are standard graph benchmarks that span two types of networks: citation and product, with different statistics. In addition, we include two additional graph datasets to test scalability of the Columnar-Embedder architecture to much larger graphs. 

\begin{table}[ht]
    \centering
    \begin{tabular}{|c|c|c|c|c|c|}
    \hline
    \hline
        \textbf{Dataset} & \textbf{N} & \textbf{Edges} & \textbf{Classes} & \textbf{Mean Degree} & \textbf{Source}\\ \hline
        Cora          & 2,485& 5,069& 7 & 4.08 & Citation network \cite{Yang2016RevisitingSL}\\ \hline
        Citeseer      & 2,120& 3,717& 6 & 3.51 & Citation network \cite{Yang2016RevisitingSL}\\ \hline
        Amazon-Photo  & 7,487& 119,081& 8 & 31.8 & Co-produce network \cite{shchur2018pitfalls}\\ \hline
        PubMed    & 19,717& 88,648& 3 & 4.5& Citation network \cite{Yang2016RevisitingSL} \\ \hline
        Co-Author Physics & 34,493& 495,924& 5 & 14.38& Coauthor Physics networks \cite{shchur2018pitfalls} \\ \hline
    \end{tabular}
    \caption{Graph Datasets Benchmark}
    \label{tab:Benchmark}
\end{table}
For each graph dataset (Cora, Citeseer, and Amazon-Photo), we evaluated the SDRs produced by Columnar-Embedder on two downstream tasks:
\begin{itemize}
    \item \textbf{Node Classification:} A 5-fold stratified cross-validation with a logistic regression classifier on raw SDRs ( or dense embeddings for the baselines) and report mean accuracy ( $\pm$ ) standard deviation across folds.  
    \item \textbf{Link Prediction:} We adopted the experiment protocol given in the n2v paper \cite{Grover2016node2vecSF}, which is conducted as follows: (1) 10-15\% of edges are held out as positive test examples, (2) an equal number of non-edges are sampled as negatives. (3) The residual graph ( after the edges are removed) remains connected. (4) The model is trained on the residual graph, and new embeddings are created. (5) Edge features are computed by four binary operators on endpoint embeddings. (6) A logistic regression classifier is trained on edge features and evaluated on held-out pairs. (7) We report AUC and AP for the best operator per method-dataset combination.
    \item \textbf{Baselines} Dense n2v (d = 64, walk length = 20, number of walks = 10, p=q=1, window = 5, epochs = 10) and 2-layer GCN (d =64) trained end-to-end with supervised labels are reported as a reference point for what state-of-the-art methods have achieved. Furthermore, we applied additional methods, including LINE \cite{LINE}, DEEPWALK \cite{DeepWalk}, and GraphSAGE \cite{GraphSage}, to the same datasets using the same protocol.
\end{itemize}
\subsubsection*{Node Classification task}
\begin{figure}[ht]
    \centering
    \includegraphics[width=0.5\linewidth]{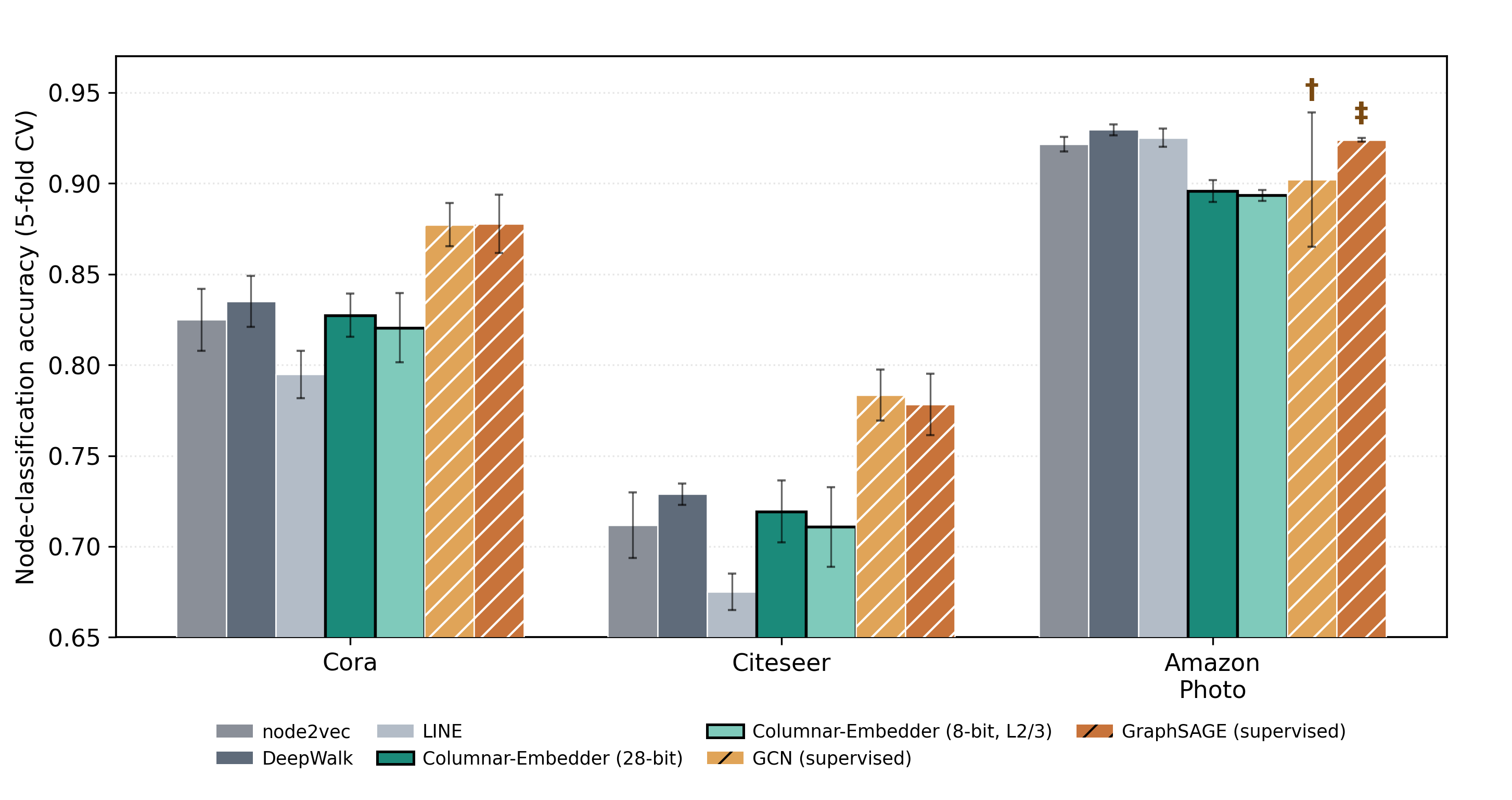}
    \caption{Node Classification accuracy on three datasets: Cora, Citeseer, and Amazon-Photo.  \textdagger GCN on Amazon-Photo is high-variance (0.902 ± 0.037). \textdaggerdbl  GraphSAGE on Amazon-Photo is bistable; a converged run (0.924) is the best run.}
    \label{Node_cla}
\end{figure}

The node classification task clearly indicates the effectiveness of the Columnar-Embedder architecture in constructing meaningful SDRs that are unique and discriminative at the class and node levels. The node classification presented in Figure \ref{Node_cla} indicates that Columnar-Embedder successfully maps the statistics of the random walk training sequences via PPMI into a binary representation of each node reflecting the neighborhood similarity. We conduct further analysis on the representation of each dataset and conclude:
\begin{itemize}
    \item The PPMI signal used in the BCM-PPMI learning rule is the precise quantity the random walk family implicitly factorizes. Thus, a learning rule that uses PPMI directly operates on the same information that gives n2v its accuracy ceiling, but via local Hebbian rules rather than skip-gram softmax. 
    \item Accuracy fluctuations across datasets indicate a greater dependence on the ability of random walks to capture structure than on the underlying assumption of uniform relations among node neighbors.
    \item Providing more auxiliary information, e.g., node features, will definitely help boost accuracies. The GNN embedder uses auxiliary information, "node features," to construct node representations. In contrast, our model does not require node features, which explains the accuracy deficit between the GNN and Columnar-Embedder architecture.  
\end{itemize}

The feature-based GNN methods of GCN and GraphSAGE struggle to outperform traditional random walk techniques because they rely on node features of a graph with a high average degree of 32, compared to the other two sparse graphs with only 3.5 and 4 node degree averages. Thus, the GNN kernel aggressively smooths node features, pushing central nodes with hub-region roles toward similar embeddings and flattening the discriminative signal, degrading the loss surface sharpness of the classifier. The Amazon Computers/Photo datasets were introduced by Shchur et al. \cite{shchur2018pitfalls} specifically to demonstrate that GNN evaluation is fragile and hyperparameter-sensitive. Therefore, instability on the Photo dataset at default settings is expected behavior, and we want to test our architecture to emphasize its resiliency.  
\subsubsection*{Link Prediction task}
The link prediction task is an important indicator of node embedding quality when the similarity of neighbors suggests that a link between them is likely to exist, and vice versa. The experiment setup is described above, and we could point out the following:
\begin{figure}[!ht]
    \centering
    \includegraphics[width=0.5\linewidth]{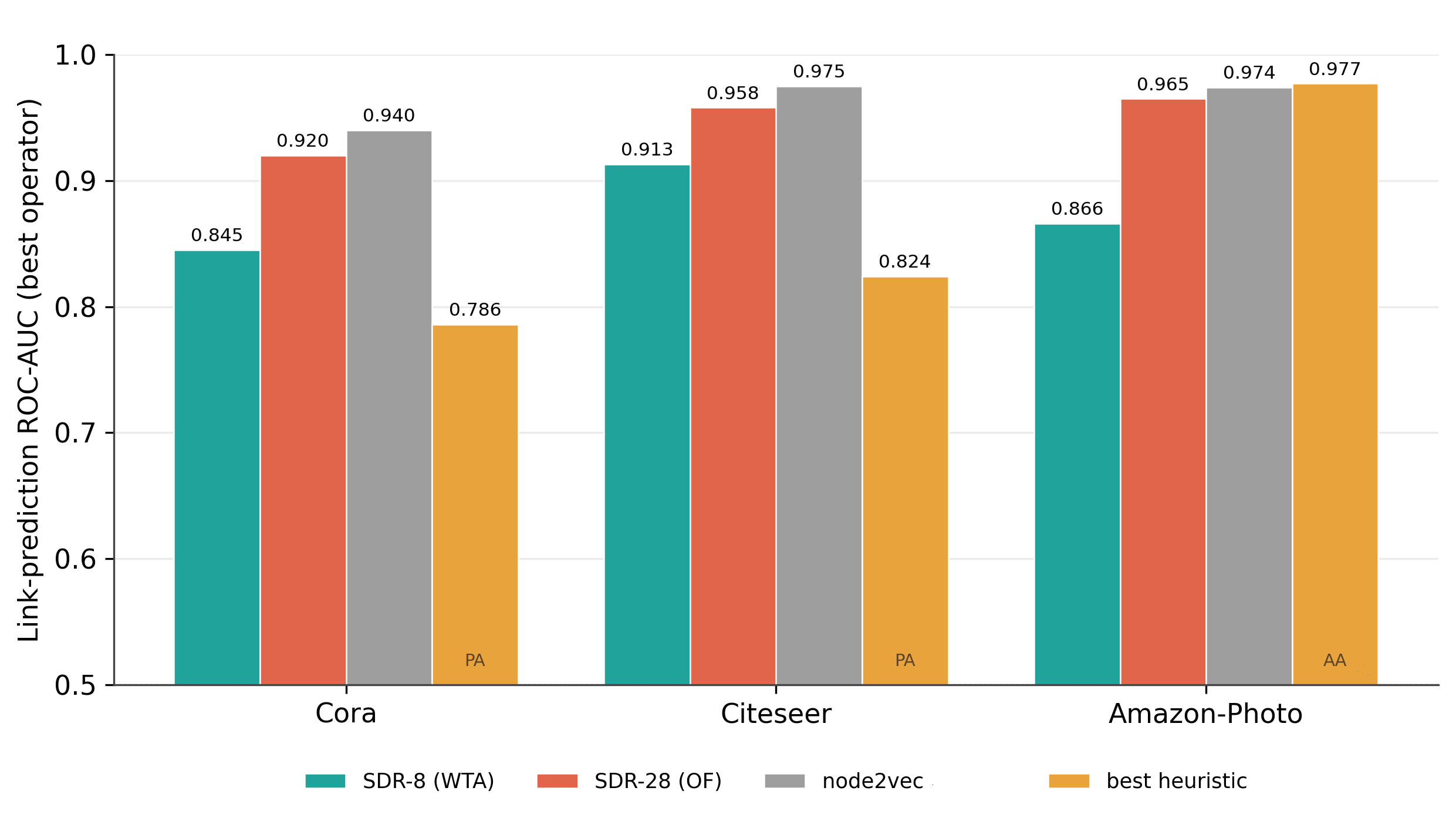}
    \caption{Link Prediction across the datasets ( Leakage-free, 15\% held out)}
    \label{fig:lp}
\end{figure}
\begin{itemize}
    \item On the residual graph, roughly $96\% $ of the held-out positive edges connect nodes that share at least one common neighbor, while $91\%$ of the random negatives share zero neighbors. The Adamic-Adar with 97.7\% accuracy could be interpreted as the answer to the question: do any two nodes have a common neighbor? For a dense graph like Amazon-Photo with an average degree of 31, real edges sit inside tight neighborhoods, and uniformly random non-edges always land between unrelated nodes. Thus, common-neighbor counting perfectly separates them.
    \item The positive test edges encounter nodes with much higher degrees (the mean endpoint degree is 86) than random negatives ($\sim 27$). That degree mismatch explains why even preferential attachment could separate those nodes. Thus, random negative sampling on a dense graph produces easy negatives. The fix for this phenomenon is actually hard negative sampling: sample non-edges that share neighbors (2-hop pairs). 
    \item The SDR is trained for node-level similarity, not specifically for edge prediction, and still achieves (92\%, 95.8\% and 96.5\%) on every benchmark, as shown in Figure \ref{fig:lp}, comparable to the baseline trained with explicit edge supervision.
    \item The SDR produced by the cortical layer SDR-8 shows that premature representation of that layer learned useful information about graph geometry, despite having only eight active neurons per node (8/1500), which is roughly $0.53\%$ sparsity. 
\end{itemize}
\subsubsection*{Operator Sweep of Columnar-Embedder on Link Prediction}
\begin{figure}[!ht]
    \centering
    \includegraphics[width=0.95\linewidth]{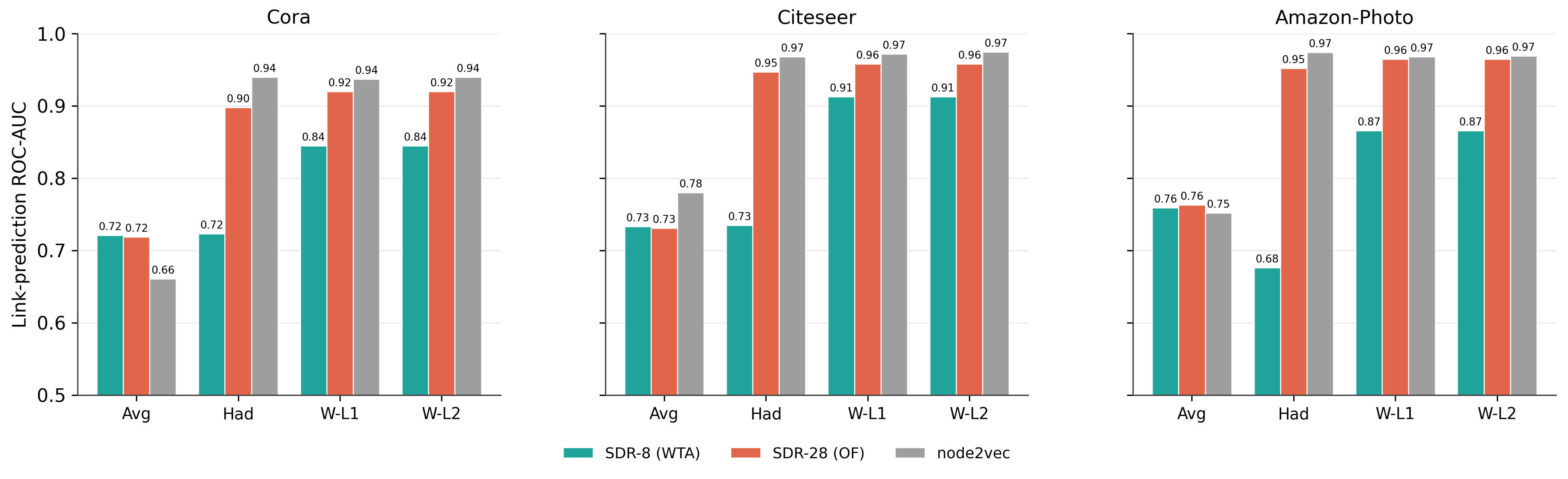}
    \caption{Operator Sweep of link Prediction task on datasets (leakage-free and 15\% held out)}
    \label{fig:sweep}
\end{figure}

Figure \ref{fig:sweep} reports the link prediction ROC-AUC on Cora, Citeseer, and Amazon-Photo for all binary operators used by each method (Average, Hadamard, Weighted-L1, Weighted-L2).  
The pattern is consistent across 3 datasets and diagnostics of how each representational scheme conveys similarity. For SDRs, Weighted-(L1, L2) are the strongest operators and are numerically identical since SDRs are basically binary features. Thus, it confirms that the predictive signal lives in the symmetric difference between sparse codes. Hadamard, by contrast, is weak for sparse SDR-8 because two 8-bit codes rarely share active bits, so their element-wise product is mostly uninformative. In contrast, SDR-28 (full representation capacity of $L_5$) recovers Hadamard performance as bit overlap increases (starts roughly at 8.7 of 28 on average). The n2v shows the opposite preference, so it is slightly favoring Hadamard because n2v is essentially a dense, continuous, real-valued representation. The average operator is the weakest overall, as averaging aims for the median, which is sensitive to outliers. Therefore, we report the SDR results under weighted-L1, which yields the best performance.      

\subsection*{Representation Quality: Sparse Binary codes vs. Dense Embeddings }
The dense n2v embedding maps each node to a 64-dimensional continuous space, whereas our model places each node in an 1800-dimensional Binary space with only 28 active bits. This striking difference makes SDR codes highly competitive with dense codes. Sparse distributed representations SDRs preserve the neighborhood geometry of dense continuous embeddings and support downstream tasks such as node classification at statistical parity, as shown in previous sections. Achieving that with only two orders of magnitude fewer active bits (28 bits vs 64×32 = 2048 stored bits ), without backpropagation, global softmax, or label supervision during representation learning, makes SDRs a reasonable choice for node representation. However, we do not claim that the SDRs outperform dense embeddings; although they are very competitive and have the potential to overtake them. The next sections provide a meaningful analysis of our representational quality, in which SDRs have clear advantages.
\subsubsection*{Geometry preservation}
The qualitative results establish that the Columnar-Embedder model recovers community structure from BCM-PPMI modulator learning alone, with no community labels presented during the training. We quantify the establishment of block structure across all three metrics: the pairwise SDR-overlap matrix, the n2v cosine matrix, and the random-walk co-occurrence matrix. 
\begin{figure}[!ht]
    \centering
    \includegraphics[width=0.65\linewidth]{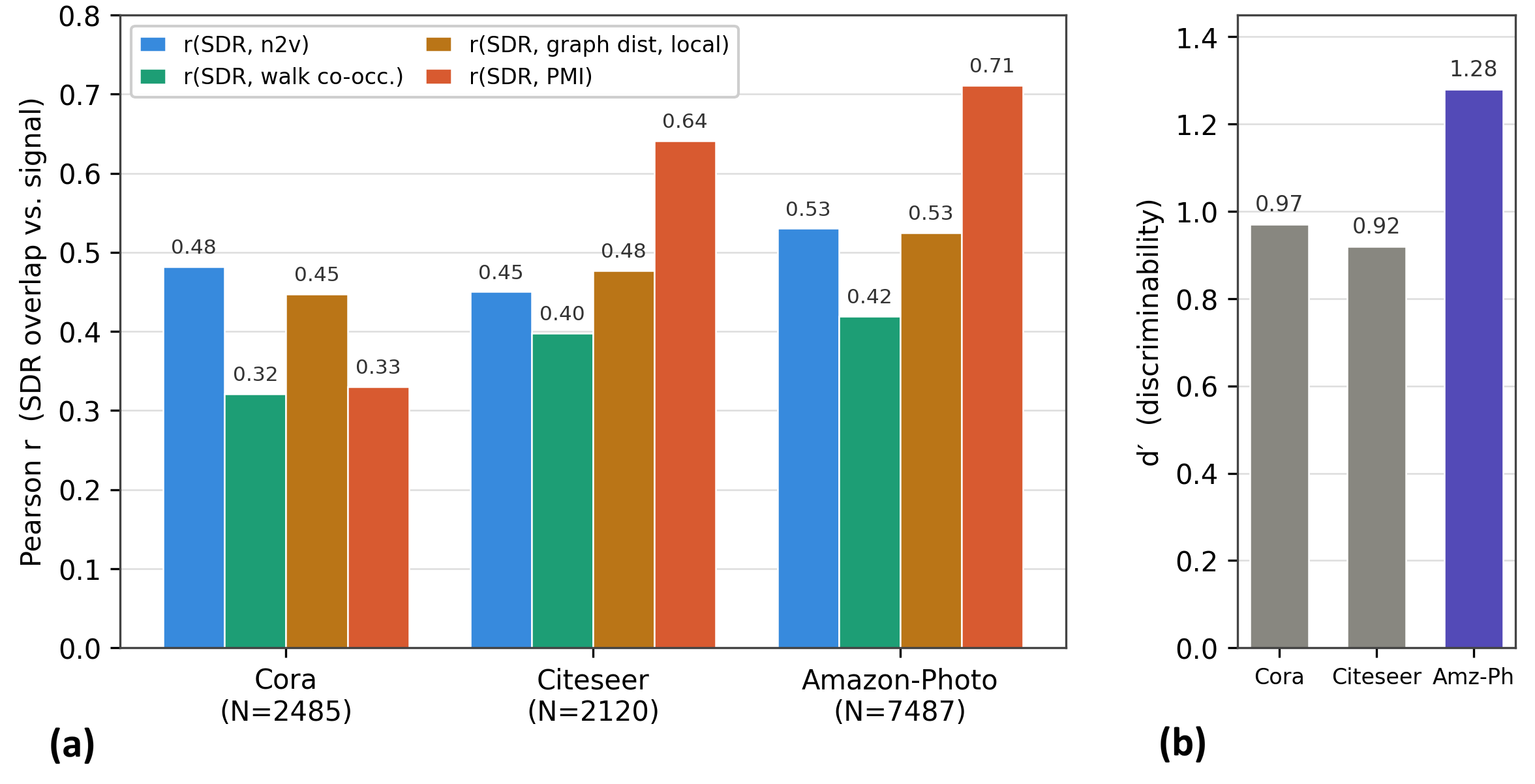}
    \caption{28-bits SDR fidelity metrics over Cora, Citeseer, and Amazon-Photo. (a) SDR correlation with n2v, walks statistics, and modulator signal. (b) shows class discriminability of SDR on benchmark datasets.}
    \label{fig:SDR_Fidelity}
\end{figure}
The fidelity metric is the correlation between the flattened off-diagonal entries of the learned SDR-overlap and those of four reference geometries, ordered from the closest to the training signal to the rawest topology. With an independent reference n2v method, Columnar-Embedder uses PPMI from walks and never sees n2v to infer the graph structure and class distribution. All values are Pearson $r$ on a 3-million-pair off-diagonal subsample, and Spearman $\rho$ is consistent in sign and ordering throughout.
\begin{figure}[!ht]
    \centering
    \includegraphics[width=0.5\linewidth]{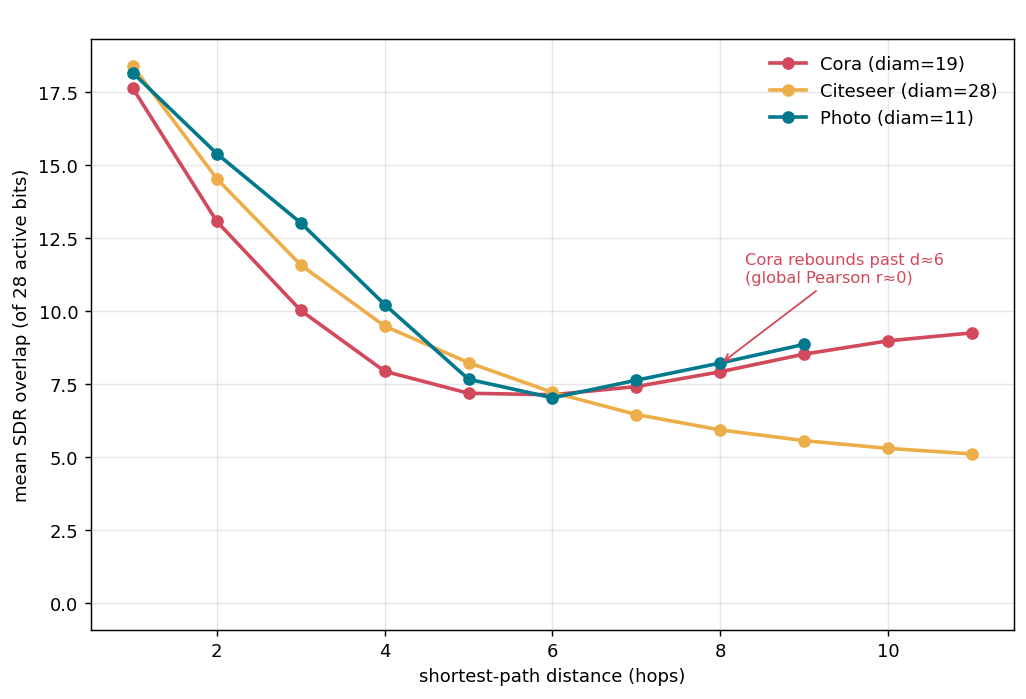}
    \caption{28-bits SDR overlap vs graph distance (hops), indicating strong geometric correlations between graph nodes' proximity and shared bits in the SDRs.}
    \label{fig:Overlap}
\end{figure}
In the first conclusion, the independent n2v agreement is positive and increases with the density of the graph. A model trained only on the PPMI signal learns the same pairwise geometry as a dense embedder, without being implicitly or explicitly shown what to learn as in Figure \ref{fig:SDR_Fidelity}. Second, the the overlap among SDRs is a local proximity measure that capture structural closeness of any pair of nodes on every graph. The graph distance column indicates that the local-regime similarity (node pairs within four hops of SDR holds, meaning that SDR is a faithful local proximity code on every graph. The overlap starts at $~18/28$ bits for directly connected pairs and decays smoothly for $d \leq 5$, which is clean and monotone across all graphs; see Figure \ref{fig:Overlap}). Third, the value $r(SDR, PPMI) = +0.33$ is notable for an informative inversion specific to the Cora data set. The interpretation is that PPMI is quantized into 300 coarse regions, whereas the learned SDRs-overlap geometry is smooth, indicating that the trained SDRs recover a continuous n2v-like structure finer than the blocky PPMI that biases its column selection. The architecture not only inherits PPMI sub-region structures, but also actively interpolates across the boundaries of the PPMI regions. The discriminability distance $d'$ across datasets is calculated as in equation \ref{discriminability}. 
\begin{equation}
\label{discriminability}
    d' = (\mu_{same-class} - \mu_{cross-class})/ \sqrt{\frac{1}{2}(\sigma_{same-class}^2 + \sigma_{cross-class}^2)}
\end{equation}
Figure \ref{fig:SDR_Fidelity} (b) describes the $d'$ measure that quantifies how cleanly the SDR geometry separates same-class node pairs from different-class node pairs based on pairwise overlap. For the Cora dataset, $d' = 0.97$, which translates to $AUC = 0.75$, so if a random same-class pair and a cross-class pair are selected, their SDR overlap is high 75\% of the time. Thus, $d' = 0$ means that the code carries no class signal, and $ d'= 1$ means that both pairs are one standard deviation apart. 

\subsubsection*{Noise robustness of SDRs representations}
\par
\begin{figure}[!ht]
    \centering
    \includegraphics[width=1\linewidth]{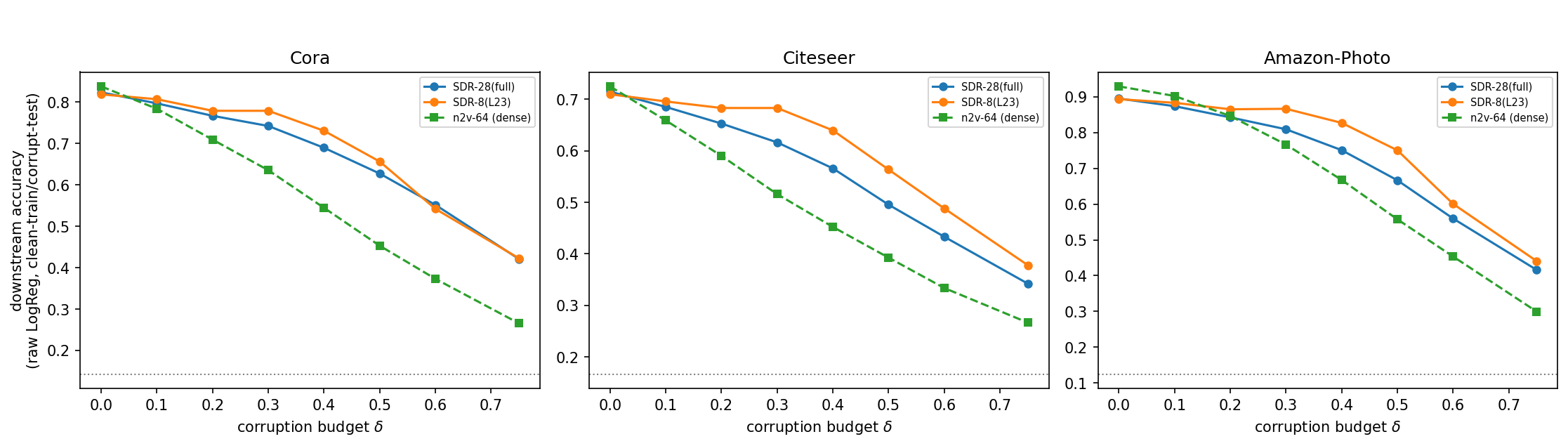}
    \caption{Robustness to corruption ( bit flipping fraction $\delta$ ) vs n2v with d =64}
    \label{fig:Robustness}
\end{figure}

 We probed the robustness of the Columnar-Embedder SDR codes to input corruption under a representation-agnostic budget $\delta = 1 - E[self-similarity]$, applied in each method's native metric. We use bit-flip for SDR, where $\delta$ denotes the fraction of bits flipped, and a closed-form per-vector Gaussian for dense n2v (d = 64). Figure \ref{fig:Robustness} shows that although n2v has retained high accurate representation on clean input (classification accuracy of 0.837/0.728/0.923), it degrades the most rapidly, so both SDR codes overtake it as corruption increases, crossing at $\delta = 0.10$ on the citation graphs and $\delta = 0.20-0.30$ on the much denser Amazon-Photo graph. In every case, well above the chance floor, with the sparsest 8-bit $L_{2/3}$ code that degrades most gracefully of all and leads n2v by $+14$,$+17$, and $+10$ points at $\delta = 0.3$. That is consistent with SDR near-orthogonal cross-class geometry.    

\begin{figure}[!ht]
    \centering
    \includegraphics[width=1\linewidth]{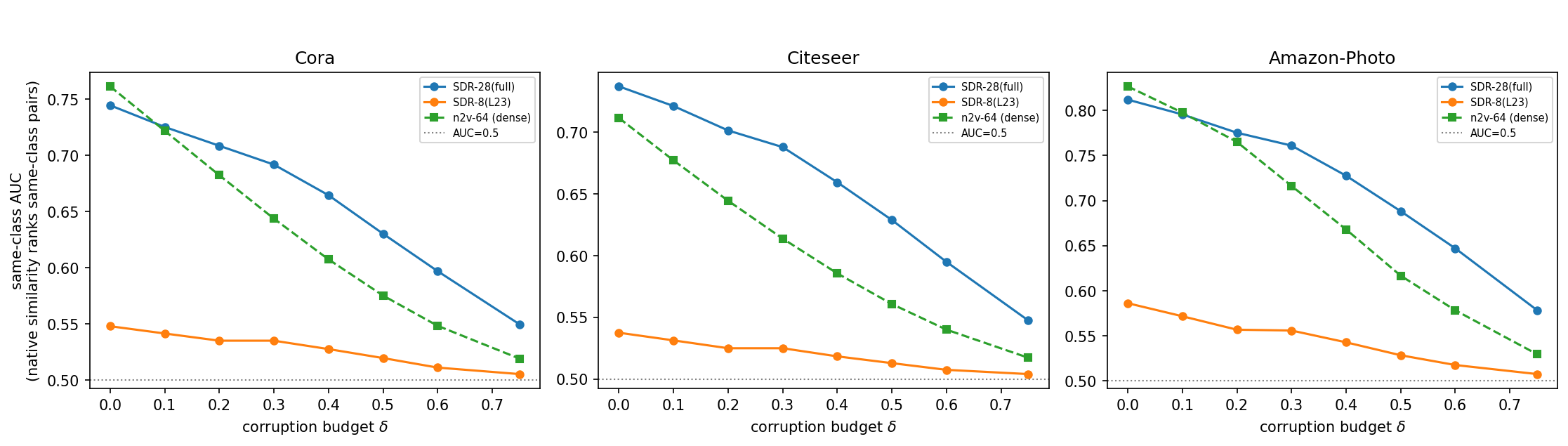}
    \caption{The class separability under corruption. Same-class AUC in each representation's native metric (SDR overlap, n2v cosine similarity)}
    \label{fig:Separability}
\end{figure}

\begin{figure}[!ht]
    \centering
    \includegraphics[width=0.5\linewidth]{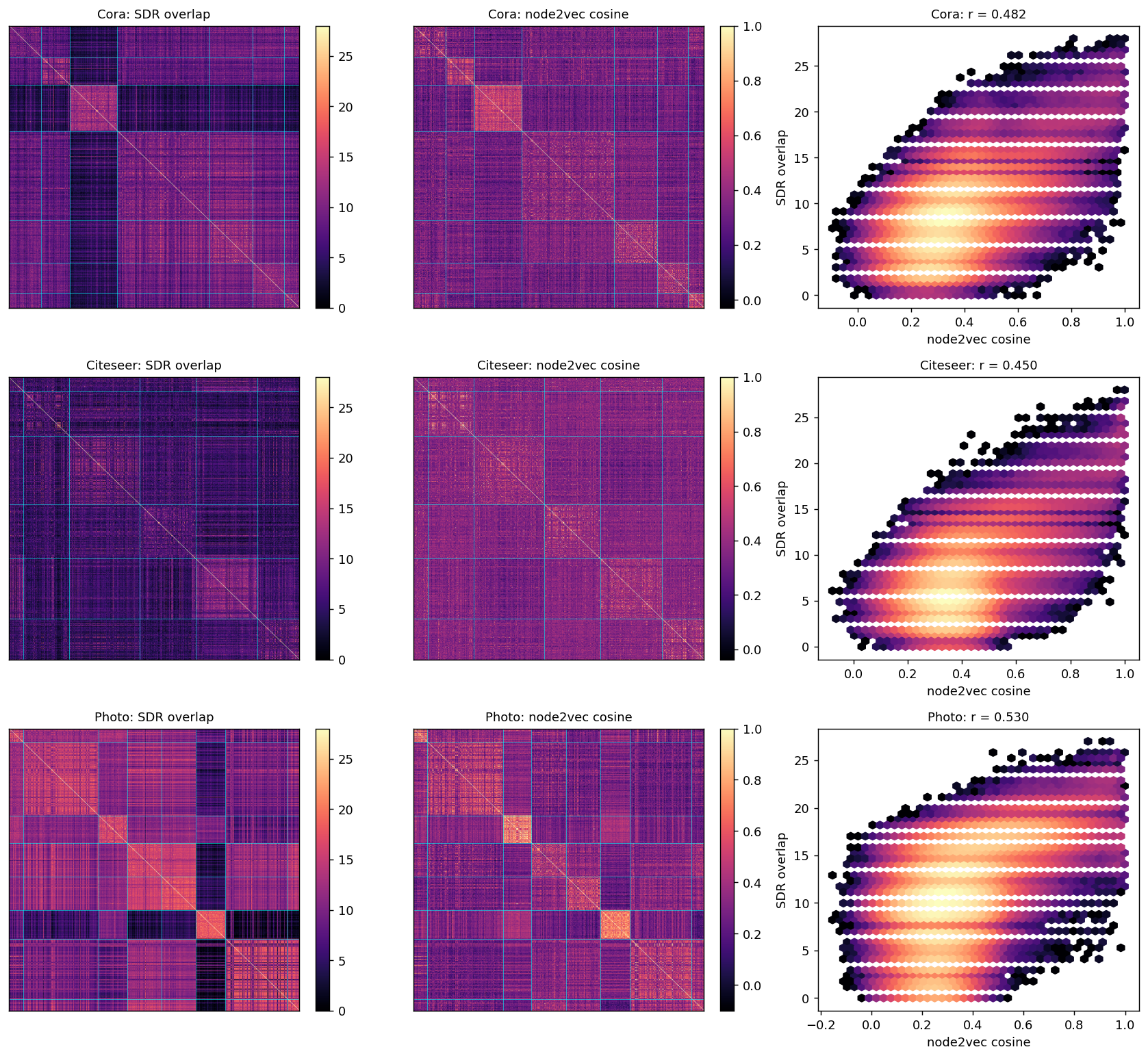}
    \caption{SDR overlap vs n2v geometry. The Similarity matrix panel shows that it sampled roughly $600$ nodes reordered by class. The hexbin R values are computed on the full 3-million-pair set. }
    \label{fig:Qayalysis}
\end{figure}

As in Figure \ref{fig:Separability}, the overlap geometry of the 8-bit and 28-bit SDRs both start near and decays far more slowly than n2v's cosine geometry, leading in all three datasets by $\delta = 0.3$. Whereas the 8-bit code's overlap-AUC remains near 0.5 throughout, which isn't a sign of fragility, because the 8-bit $L_{2/3}$ code encodes class membership by the mini-columns that fire (linearly combined) rather than by overlap magnitude. Both Figures \ref{fig:Separability} and \ref{fig:Robustness} confirm that the sparse binary cortical code trades a modest clean-input accuracy deficit for markedly shallower degradation under corruption. Therefore, the deployed 28-bit SDR wins in robust pairwise separability while the 8-bit $L_{2/3}$ code wins in robust linear decodability.  
Figure \ref{fig:Qayalysis} is the central visual evidence for the representation quality metric. For all datasets, the same block-diagonal community structure appears in both n2v and SDR representations, which clearly preserve the class geometry of dense embeddings. The Figure signifies that SDRs have the same neighborhood structure at a fraction of the active dimensionality, and the agreement strengthens from Cora to Amazon-Photo, reflecting the discriminability factor ($d'$). The above agreement indicates that the representation becomes cleaner, not noisier, as the graph provides more structural signal, all without backpropagation and without supervised labels.  

\subsection*{Architectural Resilience to noise, corruption, and missing data}
We tested the robustness of the Columnar-Embedder architecture on four different classes of perturbations. Under 5\% L4-bit flips, the K-WTA preserved column selection for $\sim 85\%$ of the nodes, with the score margin between the 8th and 9th columns typically exceeding 10\%. In addition, the $L_{5a}$ outstar readout shows a similar tolerance of $\sim 90\%$ remaining unchanged, consistent with the analytical bounds in \cite{Ahmad2016HowDN} for k-of-N codes. Replacing 10\% of the W or $W_r$ entries with zeros after training leaves both layers functional because WTA and Top-K are rank-operative rather than magnitude sensitive, confirming the degeneracy principle as in \cite{Marder2011MultipleMT}.
Plasticity learning rules are simple and efficient, so that BCM-PPMI with only 10\% of walk pairs shown recovers $\sim \geq 95\%$ of the final classification accuracy, with outstar learning rule convergence accelerated compared to before. Finally, sequential training on disjoint graphs retains 70-80\% of the first task accuracy versus 40-60\% loss for gradient-descent trained baselines. This can be explained as a result of the slow timescale of BCM's meta-plasticity threshold and IP's adaptive thresholding.
\subsection*{Architectural scalability}
 Although the results presented above show the capacity of the Columnar-Embedder architecture to scale up to $3.5 \times$ the graph size without any architectural changes or hyperparameter tuning, we went further and evaluated Columnar-Embedder on larger benchmarks: the PubMed and CoAuthor-Physics graph datasets. We followed the same protocol described in experiment protocol and conducted a node classification task to demonstrate the scalability and portability of the Columnar-Embedder architecture.  
\begin{figure}[!ht]
    \centering
    \includegraphics[width=0.5\linewidth]{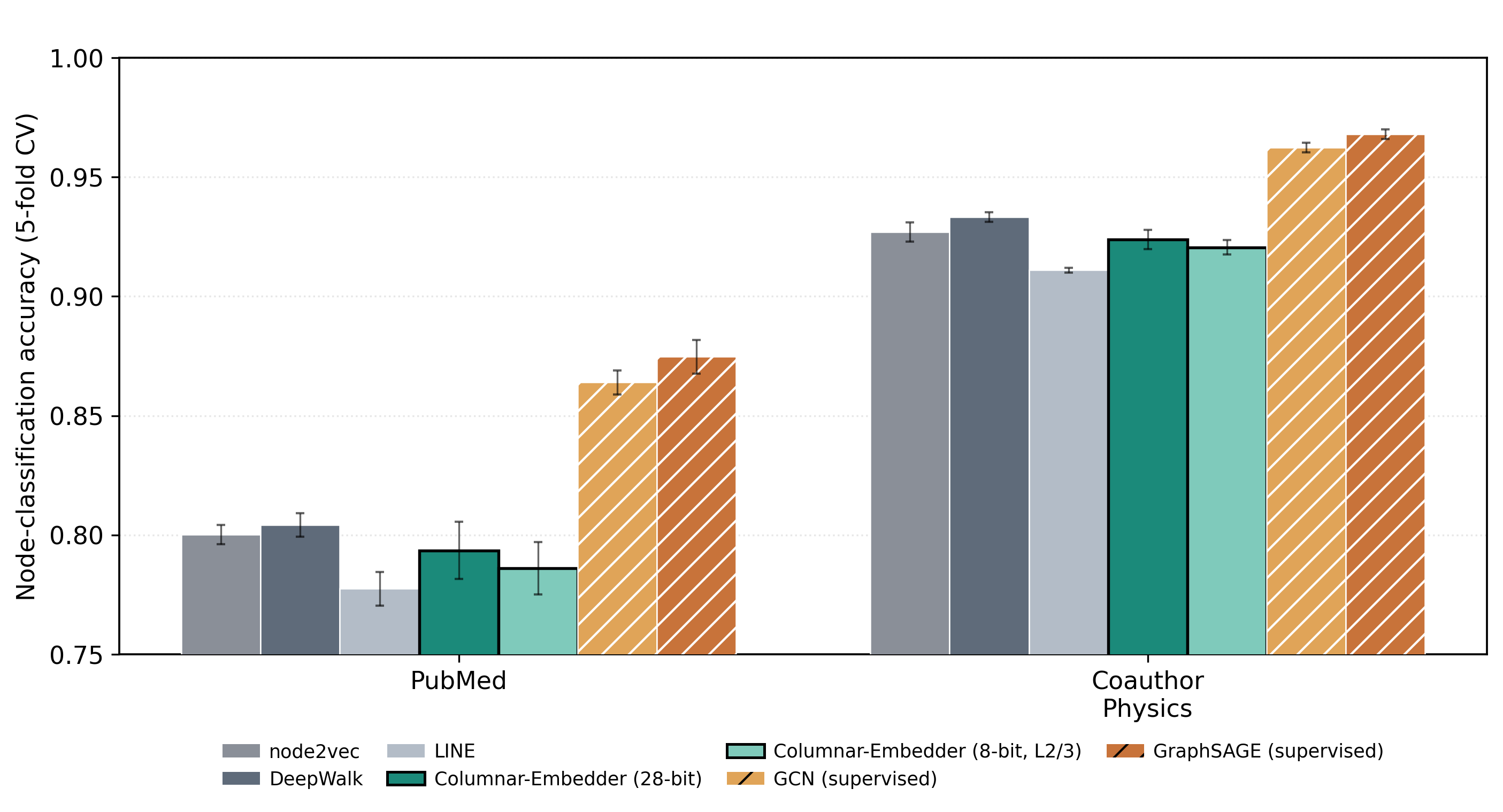}
    \caption{The Columnar-Embedder architecture scaling to larger graphs with no architectural or parameter changes. The two graphs of PubMed, and CoAuthor-Physics are 7.9×, and 13.9× the size of Cora.}
    \label{fig:largerGraphs}
\end{figure}
Figure \ref{fig:largerGraphs} offers evidence of the Columnar-Embedder architecture ability to maintain high performance even at larger scales at no SDRs quality degradation. The result of conducting the node classification task on a much larger dataset has confirmed the following points beyond the obvious competitive classification accuracy scores:

\par (1) Internal Homeostasis within the cortical columnar is scale-invariant. Both larger graphs reach essentially complete code utilization (100 \% of neurons active) and near-perfect uniqueness ($ \approx 99.9$), and both settle their within-column diversification statistic at the random-WTA floor of $1/C_{neurons}=20\%$. The same cell utilization remains near the random floor (PubMed $20.8 \%$, Physics $20.9 \%$) at $7.9 \times$ to $13.9 \times$ scales of previous Cora graph. The operating point fluctuates as observed on the dense Amazon-Photo graph ($21.2 \%$) because when the graph supplies class-coherent column structure, Anti-CE performs little prototype-collapse work and drives within-column competition to maximum diversification. The homeostatic and competitive rules thus converge to the same regime independent of graph size.

\par (2) Across the five graphs, pairwise discriminability $d'$ rises as the graph becomes structurally denser, but the trend is carried by the sparsity of the degree distribution rather than its mean. Thus, $d'$ correlates strongly with the fraction of degree 1 nodes (Spearman $ \rho \approx 0.9$) and only weakly with mean degree ($ \rho \approx 0.6$). The dataset PubMed which is the most structurally sparse graph, $46\%$ degree 1 with a median degree of 2, shows the lowest separation. The same-class node pairs share 4.7 of 28 active neurons, while the cross-class share 2.6 with $d' = 0.66$. However, denser graphs achieve wider margins (e.g, Coauthor-Physics: 7.9 vs 1.7; $d' = 1.75$). The trend is not strictly monotonic in density because Amazon Photo, the densest graph, does not reach the highest $d'$, indicating that class separability rather than raw density sets the upper limit. In every case, mixed-class collisions remain effectively absent: PubMed produces no cross-class duplicate SDRs among $19,717$ nodes, and Physics produces a single one among $34,493$ nodes, preserving the architecture's strongest safety property at scale.

\par (3) The architecture has shown robustness to graph coverage deficit and class imbalance. The two graphs probe the regimes that most threaten an unsupervised encoder, and the architecture is resilient to both. On PubMed, despite 46\% degree-one nodes and the resulting under-coverage of the walk co-occurrence matrix, the PPMI spectral statistics supply a similarity estimate for every node pair, including those never co-observed in walks so, it absorbs the coverage deficit. The residual 0.003 accuracy gap to n2v reflects the per-pair quality of the PPMI modulator rather than any deficiency in the Columnar-Embedder code, whose internal-health metrics match the Cora anchor. In Coauthor-Physics, the 50.5\% majority class does not inflate the classification accuracy (0.924) which far exceeds the majority-class baseline (0.505). On the other hand, macro-$F_1 of 0.898$ confirms balanced separation, and the smallest class ($2,753$ nodes) attains $F1 = 0.894$, which is stronger than a larger class. Thus, the minority-class degradation seen on the sparse Citeseer graph does not recur.
\section*{Discussion}
\par The final sparse binary node representation of the Columnar-Embedder inherits the content-addressable attractor structure of Hopfield \cite{Hopfield1982}, Willshaw et al. \cite{Willshaw1969NonHolographicAM}, and BCPNN \cite{Lundqvist2006}, which natively contrasts with dense gradient-trained embeddings. Thus, SDRs offer Pattern completion, capacity-bounded memory, and noise robustness because they are structural properties of the sparse encoding rather than learned behaviors. The architecture performs a constrained spectral decomposition in the PPMI-derived subspace. Thus, BCM learning rule locates the spectral subspace that matters for explaining the current input, while Anti-CE and IP choose a sparse, mini-column-structured basis with it. This bridges three traditional research areas: BCM/Oja as a PCA learner \cite{oja-simplified-neuron-model-1982, Gerstner_Kistler_Naud_Paninski_2014}, spectral graph theory \cite{belkin2001laplacian, HOPE}, and PPMI-factorization \cite{Qiu2017NetworkEA}. Furthermore, this connection of a deflated BCM/Oja rule could be deployed in the form of the Generalized Hebbian Algorithm  GHA (Sanger's rule \cite{Sanger1989OptimalUL}) followed by a self-organizing map (SOM \cite{Kohonen1982Self}) layer to ensure the convergence to the same top-k subspace with no eigensolver, or batch operation. We are unaware of any prior work that connects local Hebbian plasticity to graph spectral decomposition via PPMI in exactly this way. Since cortical weights $W$ are fully trained via BCM-PPMI, they concentrate $84 - 90\% $ of their energy in the PPMI's top-300 subspace, suggesting that BCM with PPMI modulation can be characterized as constrained spectral graph decomposition. 
\par A fully self-contained Columnar-Embedder architecture makes the model compatible with online and stream graph setup (dynamic graph embedding). Replacing the current LIF spiking forward pass with a more dedicated spiking variant becomes straightforward because PPMI is a natural neuromodulatory signal in the three-factor STDP frameworks. The Columnar-Embedder architecture achieves very competitive accuracy in node classification and link prediction tasks across three datasets without changing any hyperparameters other than the PPMI calculation, which depends on the graph being embedded. To the best of our knowledge, this is the first demonstration that a no-backpropagation, no-label, biologically inspired local plasticity architecture transfers without tuning across both citation and co-production graphs over the specified ranges of size and density.  
\par Handling different graph sizes demonstrates that internal neuron-level Homeostasis is scale-invariant. Scaling up the graph size did not degrade neuron utilization, with almost $100\% $ of neurons active and participating in cortical activities. The SDR representation has enormous advantages over the dense n2v representation with d=64 in data compression, query speed, and energy consumption during training and inference. For instance, on Cora, SDR has $2.3 \times$ lower memory requirements under the bitmap compression regime, a $15 \times$ memory advantage (reduced memory requirements) with the sparse index implementation, and a $23 \times$ memory advantage with the entropy-coded implementation. At small scale graphs like the Cora  (2485 nodes), it is not a significant encoding saving. However, with a 1 Million-node graph, a sparse-index representation fits in 34 MB versus 512 MB for float32, which allow the smaller data structures to fit in the smaller and faster caches.
\par Our model generalizes to direct/indirect, featureless/featured, and homogeneous graphs, and scales to larger graphs without architectural changes. The robustness reported above is empirical and specific to the perturbations we tested; nevertheless, our architecture is still brittle to (i) adversarial $L_4$  pattern engineered to flip column ranking, (ii) graph topology shifts that change the spectral regions used to build eligibility.  (iii) very small training sets where the Outstar rule has insufficient pattern exposure to develop selective $L_5$  receptive fields or where PPMI estimates from walks are too noisy to be useful. However, in future work, we will develop alternatives that mitigate these vulnerabilities.
Finally, the Columnar-Embedder architecture is a genuine attempt to utilize a cortical structure with an underlying learning mechanism to facilitate learning graph structure by constructing a compact binary representation of graph entities. The modular design of the Columnar-Embedder architecture points toward new avenues of neuromorphic implementations \cite{11196167}, dynamic graph embeddings, Query answering over graphs, and autoencoder denoising tasks.     
\section*{Methods}
\subsection*{The model Architecture}
\label{Architecture}
\begin{figure}[ht]
    \centering
    \includegraphics[width=0.65\linewidth]{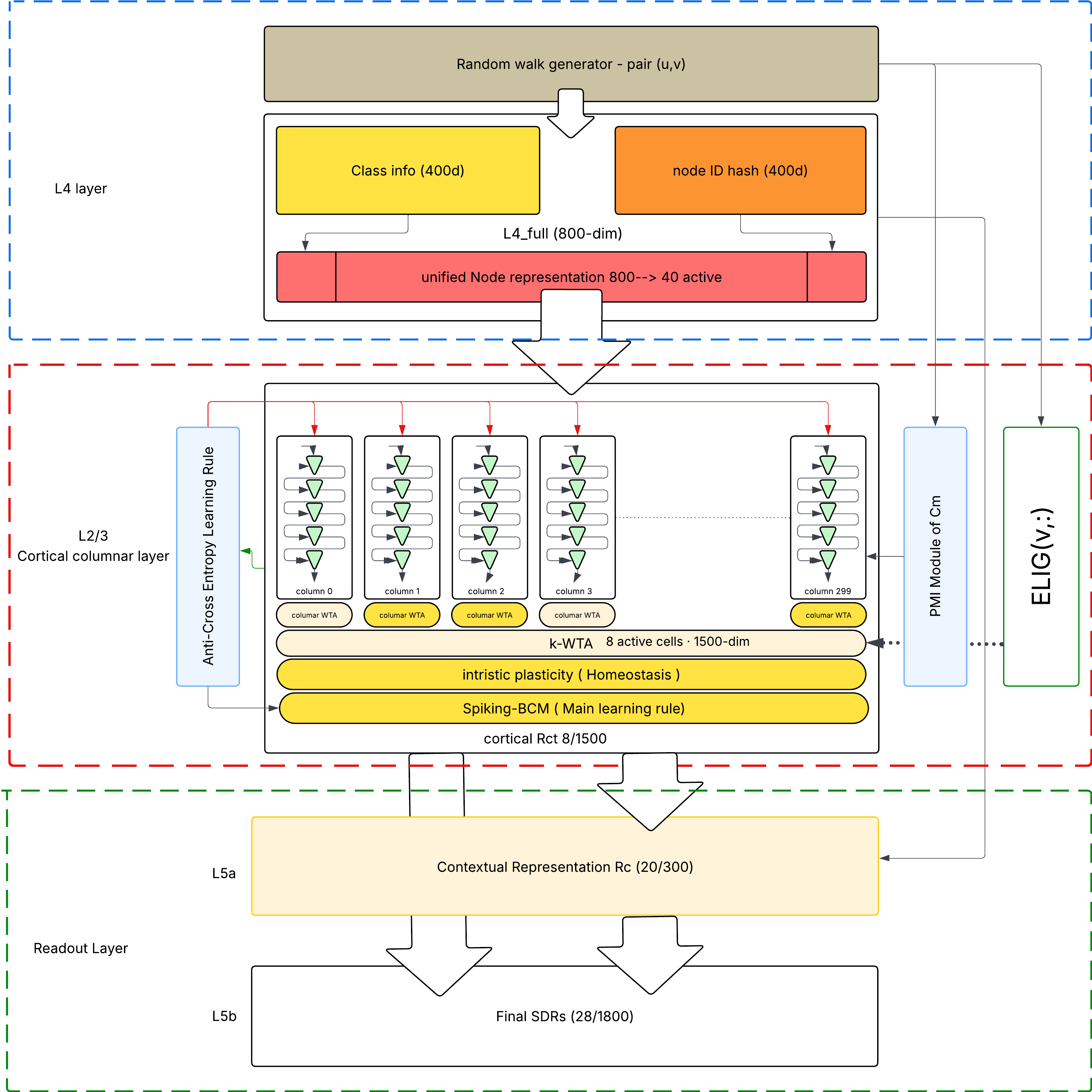}
    \caption{The proposed feedforward Columnar-Embedder Architecture of three main layers: an input encoding layer, cortical columnars layer, and Readout/Olfactory layer. }
    \label{fig:ColumnarEmbedder}
\end{figure}

The overall pipeline of the Columnar-Embedder architecture is a multi-stage processing unit in which each layer specializes in a dedicated task: encoding, processing, and projection. Figure \ref{fig:ColumnarEmbedder} shows an abstract layout of our proposed cortical-based embedding architecture, in which the computational flow starts with sequences of random walks over a graph and ends with projected SDRs in the output Binary space. Our architecture adopts these key aspects of cortical structural organization\cite{BCPNN, Hawkins2017ATO, Olfactory1} that are particularly relevant here: cortical mini-columns serve as discrete units of attractor memory, and learning is driven by log-probability ratios of pre- and post-synaptic activity. 
The model uses two separate information pathways in order to map the co-occurrence statistics of pairs of nodes ( as a proximity measure on the graph side) captured by random walks over the graph to proportional representational similarity (a similarity measure on the final SDR side). In the first path, random walk sequences are presented to the encoding layer $L_4$, which encodes the temporal aspect of walk transitions of the current training sample. This feedforward path maps similar nodes into similar cortical regions in the $L_{2/3}$ columnar layer by amplifying the target-column drive based on the current walk step. Second, the co-occurrence statistics matrix PPMI is used to perform all learning steps in cooperation with BCM to decide correlation strengthening between active columns based not only on the current step but also based on previous instances of the same pair without feedback, labels, or backpropagation. Next, we will describe in detail all components of the Columnar-Embedder architecture.       
\subsection*{Layer L4: Granular input encoding}
\label{L4}
The layer consists of 800 neurons decomposed into two groups: $L_4$ class and $L_4$ ID.
\begin{itemize}
    \item The $L_4$ class: is a group of 400 neurons in which only 20 are active per node. Their main function is to aggregate the identities of the neighbors via $A \cdot R$, followed by a top-K selection per node. Where $A$ is the adjacency matrix, and R is a random binary projection. This scheme produces a class-discriminative input pattern in which nodes belonging to the same class exhibit substantially overlapping $L_4$  class activity.
    \item The $L_4$ ID is a group of 400 neurons with 20 neurons active per node as above. However, the $L_4$  ID is a random binary hash of the node index. Thus, it ensures that every node has a distinct pattern because these hashes produce identity-orthogonal codes.   
\end{itemize}
The two subgroups of representational codes are concatenated to form an 800-neuron vector with 40 active neurons per node, shown in random walks. This combination balances the class discriminative that BCM exploits to learn class-coherent columns with identity disambiguation, which Anti-CE and the Outstar readout exploit to maintain uniqueness across same-class nodes. The split mirrors the olfactory bulb's glomerular code where combinatorial sparse activity is generated by overlaying a topographically informative signal with a deliberately non-topographic random signal.   
\subsection*{Layer L2/3 cortical columnars}
\label{L2/3}
The 1500-neuron $L_{2/3}$ layer is arranged as 300 mini-columns \cite{Hawkins2017ATO} of 5 neurons each. Each neuron receives an excitatory drive from $L_4$ through the plastic weight matrix $W \in \mathbb{R}^{800 \times 1500}$. Within the column, the neurons engage in first-to-fire WTA, which in turn is driven by the integration step of Leaky-integrate-and-Fire neurons with shared inhibition. In addition, across mini-columns, the learned anti-Hebbian matrix of size $L_{col} \in \mathbb{R}^{300 \times 300}$ implements graded lateral inhibition, selecting 8 columns per input node. The result of dual-WTA is a fixed sparsity of $\frac{8}{1500} = 0.53\% $ that adapts to the varying nature of the input nodes. The choice of WTA as the within-column nonlinearity is principled rather than expedient. Maass \cite{Maass2000OnTC} proved that a single k-WTA applied to the weighted sum of inputs can compute any Boolean function, and a soft k-WTA can approximate any continuous function using only positive weights under the linear sum. This approximation matches the cortical asymmetry between excitatory and inhibitory connections, where only $\sim$ 15\% of cortical synapses are inhibitory. That shows that our use of nonspecific lateral inhibition does not sacrifice computational universality.
\subsection*{L5 Readout layer / Outstar Olfactory layer}
\label{L5}
The readout layer $L_5$ has two sub-populations: $L_{5a}$ consists of 300 neurons that receive projections from $L_{2/3}$ through a learned weight matrix $W_r \in \mathbb{R}^{300 \times 1500}$. During inference, the activity of each node's $L_{5a}$  is computed by ranking columns by $L_{2/3}$ drive and selecting the top $K_{of} = 20$ runner-up columns, i.e., those ranked 9-28 that drive the 300-neuron of $L_{5a}$ via $W_r$. The layer $L_{5a}$ is a sparse expansion read of the $L_{2/3}$ code by a Grossberg Outstar rule \cite{Arbib1970OnTC}, which is an analog of the piriform/Kenyon neuron stage in the olfactory and mushroom body circuits. The main role of $L_{5a}$ is to increase the combinatorial capacity of $L_5$ alone, thereby ensuring sufficient disambiguation of residual duplicate clusters left behind by $L_{2/3}$.
On the other hand, layer $L_{5b}$  serves as a union space for both $L_{2/3}$ sparse codes and $L_{5a}$, concatenating them to produce final SDRs of $28/1800$, which is roughly $1.56\%$ sparsity. 
The construction of $L_{5a}$ is analogous to the mammalian piriform cortex and the insect mushroom body, which share an architectural motif of a sparse high-dimensional expansion layer (piriform pyramidal neurons in mammals, Kenyon neurons in insects) that receives projections from a much smaller upstream layer (mitral neurons in the olfactory bulb, projection neurons in the antennal lobe). Such an architectural construction produces decorrelated representations through random or learned projections combined with sparse activity \cite{bazhenov2010forward, Stettler2009RepresentationsOO}. A minority of projection neurons ($\sim$ 50 in Drosophila) project to a much larger number of Kenyon neurons ( $\sim$ 2500), of which only a few percent are active for any given stimulus. The difference here is that the projection is non-topographic and learned rather than random, which allows Kenyon neurons' connectivity to be modified by experience-dependent plasticity during the training. The $L_{5a}$ (Outstar-dominated) implements a sparse expansion from 1500 to 300 neurons with selective connectivity learned by the Grossberg Outstar rule driven by post-trained $L_{2/3}$ SDR rather than by the raw $L_4$ input. The choice of the number of expansion neurons $K_of = 20$ with ( $\approx 6.7 \%$) sparsity in $L_{5a}$ is biologically motivated by Kenyon neurons in the mushroom body that spike at a steady rate for any given odor \cite{PrezOrive2002OscillationsAS}. Practically, It is also empirically robust because $K_of$ in the range of 10-40 produces 99-100\% uniqueness at $< 1pp$ accuracy variance. 
\subsection*{The five local plasticity rules}
All five learning rules that we adopted are strictly local, simplified to accelerate computation, and depend on the quantities available at that synapse or neuron. In addition, we design these rules and their modulators (the PPMI scalar for BCM and the column mask for Anti-CE) so that computation requires information derived from random walks at no extra cost. In this sense, none of the rules requires backpropagation, gradient through time, a separate feedback pathway, or any of the auxiliary machinery used by biologically plausible deep-learning alternatives \cite{zhu2022spiking}.
\begin{table}[htbp]
    \centering
\begin{tabularx}{\textwidth}{|X|l|X|X|X|}
    \hline
        Rule & Update & $\eta$ & operates on & Role \\  \hline \hline
        BCM (PPMI-modulated) & $\Delta W \propto \eta_{BCM} \cdot L_{4}^\intercal \cdot r_{post} \cdot (r_{post} - \theta_{M}) \cdot m_{PPMI}(v,u) $ & 0.010 & $W (800 \times 1500)$ & Class-coherent column selection (Accuracy)\\ \hline
        Anti-CE & $\Delta W = \eta_{ACE} \cdot L_4^\intercal \cdot (S_{5a}^{soft} - L_{2/3}) \cdot col_{mask}$ & 0.002 & Active columns of  W(:,i) & Within-columns neuron diversification (uniqueness at $L_{2/3}$)\\ \hline
        Anti-Hebbian (lateral inhibition) & $\Delta L_{col}(j,i) \propto - \eta \cdot col_{active}(i) \cdot col_{active}(j)$ & 0.005 & $L_{col} \in \mathbb{R}^{(300 \times 300)}$ & Sparse column selection ($M_{active} = 8 $ winners)\\ \hline
        Intrinsic Plasticity (IP) & $\Delta V_{th}(c) \propto \eta_{IP} \cdot (\bar{r}_{rate}(c) - r_{target})$ & 0.05 & Per $L_{2/3}$ neuron & Homeostatic threshold stability\\ \hline
        Outstar (Grossberg rule) & $\Delta W_r(j,i) = \eta_{OF} \cdot L_{5a}(j) \cdot (L_{2/3}(i) - W_r(j,i))$ & 0.01 & $W_r \in \mathbb{R}^{(300 \times 1500)}$ & $L_{2/3}$ -> $L_{5a}$ Olfactory readout. \\ \hline
\end{tabularx}
 \caption{Columnar-Embedder learning Rules}
     \label{rules}
\end{table}
In the Anti-CE row, $S_{5a}^{soft} = col_{softmax}(L_4 \times W)$ is an internal real-valued column softmax used only to compute the Anti-CE error signal. Thus, it shares the existing weight matrix W and is not a separate cortical layer. This is distinct from the binary $L_{5a}$ Outstar Readout described in the section above, which has its own weight matrix $W_r$ and is trained by the Outstar rule. The Spike-based BCM is a version of BCM that depends on LIF neuron spike times where the rate is defined as the LIF firing-rate proxy: 
\begin{equation}
    r_{post}(c_i) = exp(-(T - t_{spike}) / \tau_{LP})
\end{equation}
For spiked neurons $c_i$, otherwise 0 and $\tau_{LP} = 10ms$ and $\theta_M(c_i)$ tracks running mean of $r_{post}(c_i)^2$ at $\eta = 0.01$, and T is the forward pass computational steps. 
\subsubsection*{PPMI-Modulated BCM}
\label{PPMI-BCM}
The three-factor BCM rule modulates each synaptic update by a scalar $m_{(u,v)}$ computed for the current training pair (u,v). Our model introduces a novel technique for computing $m_{(u,v)}$ from raw walk statistics, without relying on any other embedding methods. 
\begin{equation}
    m_{PPMI}{(u,v)} = min \big  (  M_{max}, max \big ( 0, log \frac{C(u,v) \cdot T}{C(u).C(v)}\big ) \big )
\end{equation}
Where $C(u,v)$ is the current count of co-occurrences of u and v within a window of 5 along random walks, $C(u)$ and $C(v)$ are marginal counts, $T$ is the total pair count, and $M_{max} = 2$ clips runaway log-ratios. This is positive pointwise mutual information $PPMI(u,v)$, which the same random-walk methods implicitly factorize \cite{Qiu2017NetworkEA, Levy2014NeuralWE}. Four reasons that motivate this choice:
\begin{itemize}
    \item \textbf{Self-contained pipeline}: PPMI is computed directly from raw online walks without pretrained external embeddings, making our architecture end-to-end learnable from the adjacency alone.
    \item \textbf{Information-theoretic foundation:} PPMI is the quantity that random-walk methods implicitly factorize. Our model reaches the same theoretical target through local Hebbian learning rather than matrix factorization.
    \item \textbf{BCPNN alignment:} PPMI is the natural cortical learning signal in the BCPNN family \cite{johansson2005mean}. Our model uses BCM with PPMI, which mimics the BCPNN log-probability ratio rule.
    \item \textbf{Biological plausibility via eligibility traces: } PPMI is computable online from co-occurrence counts, exactly the quantity that Hebbian eligibility traces accumulate, with no backpropagation or supervisory signal required.
\end{itemize}

\subsubsection*{Anti-CE within-column diversification rule}
\label{Anti-CE}
Our model implements a clean bisection of representational responsibility in the $L_{2/3}$layer. The BCM rule determines which column wins among the 300 mini-columns that fire for each node in the encoding layer, thereby enforcing class-coherent column selection. On the other hand, anti-cross-entropy determines which of the 5 neurons within each active column wins, reflecting node-specific neuron selection. For each training node $v$, the forward pass computes the $L_4$-to-$L_{2/3}$ drive $D = L_4^{full}(v) \cdot W \in \mathbb{R}^{1500}$, partitioned into 300 mini-columns where two readouts of the same $D$ are then computed within each column $c$:
\begin{align}
     L_{2/3}(c,k) = \mathbbm{1}(k= \arg\max_{k'}D[c,k'])    &&  \text{(hard binary 1 of 5 neurons per column)}
\end{align}
\begin{align}
     S_{5a}^{soft}(c,k) = \frac{exp\big( D[c,k] \big)}{\sum_{k'=1}^{5} exp\big(D[c,k']\big)}      && \text{(soft probability over 5 neurons, sum to 1)}
\end{align}   
$L_{2/3}$ is the binary code consumed downstream of \textbf{$L_{5b}$} to be concatenated with \textbf{$L_{5a}$}. $S_{5a}^{soft}$ is a real-valued internal prediction that exists only to provide the anti-CE error signal. The pairing of a hard binary winner and a soft probability over the same drive is structurally analogous to the temporal-WTA construction of Oster et al. \cite{Oster2009ComputationWS} and to Maass \cite{Maass2000OnTC} soft k-WTA work. The error signal is computed as:
\begin{equation}
    \epsilon[c,k] = L_{5a}^{soft}[c,k] - L_{2/3}[c,k].
\end{equation}
Referring to Table \ref{rules}, $col_{mask}$ is the 1500-bit indicator of which neurons belong to active columns. The comprehensive rule list describes how the update weakens the synaptic weights of the winning neuron  (negative $\epsilon$ at $k^*$) and strengthens the weights of the four losing neurons (positive $\epsilon$ at $k \neq k^*$). Thus, the next time the same input pattern arrives, the winner is slightly less reliable, making one of the previous losers slightly more competitive. During training, the within-column WTA becomes increasingly indifferent among neurons, and different nodes recruit different neurons within the same column, thus introducing diversification in columnar activity and increasing uniqueness.
In conclusion, with $C_{neurons} = 5$ neurons per column, the same-neuron co-occurrence rate has a hard lower bound of $1/C_{neurons} = 20\%$ even with perfectly uniform random within-column WTA. Thus, two nodes of the same class will share a neuron in any given column with probability $1/5$. This issue caused our model to produce duplicates in SDRs for some nodes belonging to the same class. Therefore, the Anti-CE mechanism drives the empirical same-neuron rate toward the theoretical 20\% floor.  
\subsubsection*{Grossberg Outstar learning rule}
\label{Outstar}
The Grossberg Outstar update as in Eq. \ref{Outstar_eq} pulls $W_r$'s row $j$ toward the current $L_{2/3}$ pattern whenever $L_{5a} $ neuron $j$ is active.
 \begin{equation}
 \label{Outstar_eq}
     \Delta W_r[j,i] = \eta_{OF} \cdot L_{5a}[j] \cdot \big(  L_{2/3}[i] - W_{r}[j,i]\big)
 \end{equation}
Over many node presentations, each $L_{5a}$ neuron becomes selective for a specific cluster of $L_{2/3}$ patterns. The $K_{of} = 20$ active "runner-up" readout columns are ranked 9-28 by drive, rather than the top 8 that produced $L_{2/3}$, thereby disambiguating any residual duplicate $L_{2/3}$ that might have occurred. This is the structural complement of the Anti-CE diversification rule in which both mechanisms handle duplicates in the SDR space. The rule is information-preserving by construction because a perfect Outstar rule with orthogonal $W_r$ would reproduce the $L_{2/3}$ pattern exactly, which allows us to concatenate both $L_{2/3}$ and $L_{5a}$ representations $[L_{2/3} || L_{5a}] \rightarrow L_{5b}$ safely rather than substitute for each other. The final SDR is the concatenation of  $[L_{2/3} || L_{5a}]$ to $L_{5b}$ of 1800 bits with only 28 bits active.
\begin{figure}[ht]
    \centering
    \includegraphics[width=0.75\linewidth]{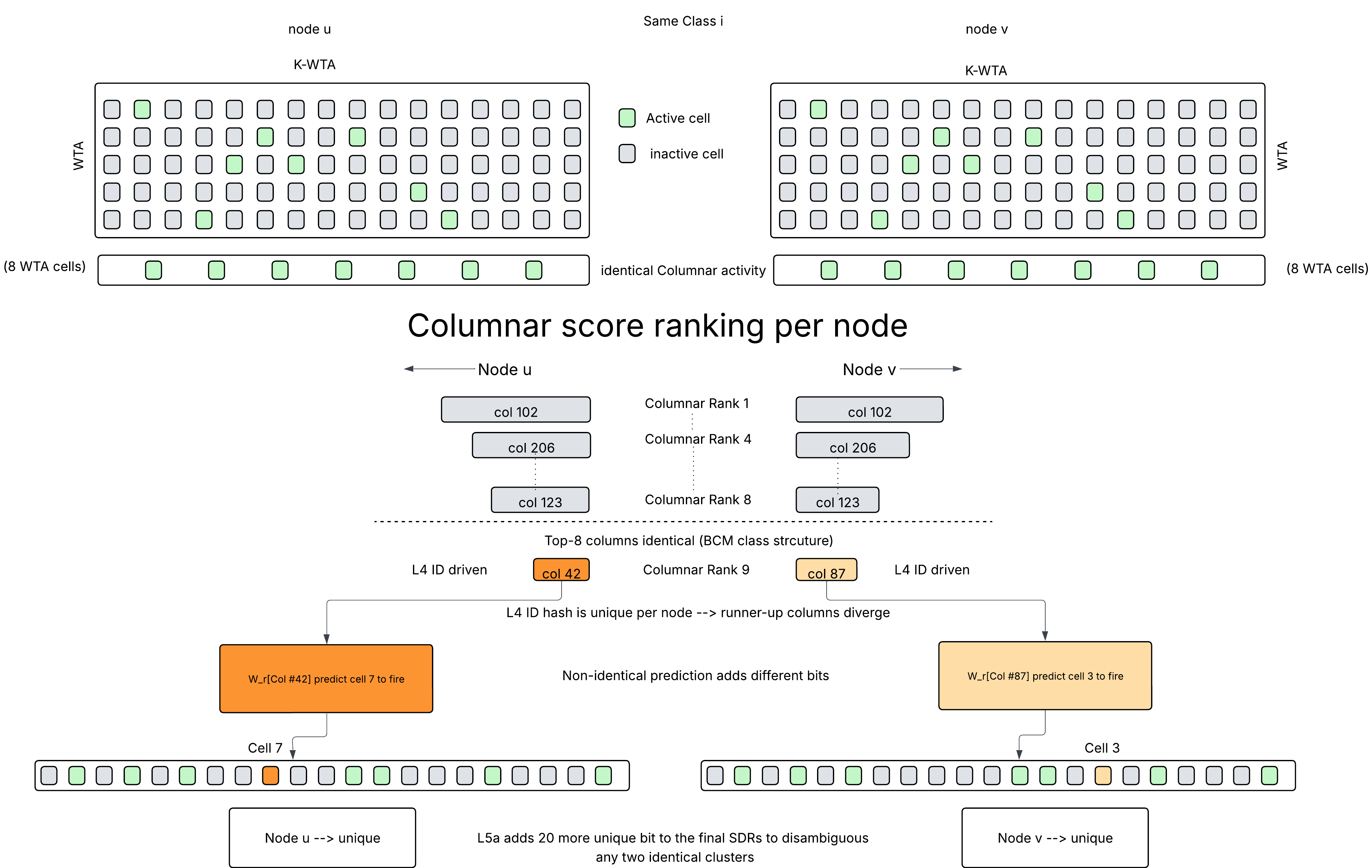}
    \caption{SDRs disambiguation process of two identical clusters representation of nodes at L2/3. }
    \label{fig:SDRs disambigous}
\end{figure}
After the unsupervised stage of the BCM-PPMI, IP, and Anti-CE stage, the eight winner-take-all $L_{2/3}$ neurons encode class-coherent structure. However, there is a possibility of a collision between two nodes that share the same 1- and 2-hop neighbors. That collision results in exact SDRs with different nodes. Figure \ref{fig:SDRs disambigous} explains the collaborative effort of all four Hebbian learning rules to disambiguate the SDRs of two structurally similar nodes. The $L_{5a}$ solves the duplication issue by construction, so any two nodes $(u,v)$ that get 8 identical cortical columnar winners and share the same-class coherent structure force BCM to focus on shared class structure rather than node identity. Duplicate elimination exploits the fact that this degeneracy breaks at the runner-up columns ranked between 9-28 because each node has a unique activity score ordering so that $v$ and $u$ select different sub-threshold columns (e.g., col \# 42 versus col \# 87). The outstar readout matrix $W_r$ then maps each node's divergent runner-up column to a distinct predicted neuron (col \#42 activates neuron \#7, and col \#87 activates neuron \#3). These predictions are projected into $L_{5b}$ and concatenated with downstream SDRs from the cortical columnar to form final SDRs. This resolves all colliding nodes and drives SDR uniqueness toward 100\% without backpropagation, label supervision, or any global tie-breaking heuristic.
\subsection*{The Helpers: Nrw, Cra, and Eligibility}
The Columnar-Embedder architecture employs a winner-take-all mechanism implemented on two levels. A columnar level where cortical columnar compete to express the current node representation to best match the most expressive neighborhood cluster. neuronal level where WTA is implemented via first to fire selection to ensure representational diversity and eliminating duplicate representation for nodes with identical 1-hop neighbors. 
The core idea is to use PPMI co-occurrence matrix to help bias WTA on the columnar level to ensure similar nodes (belong to the same cluster or label) in the original graph mapped into similar cortical regions ( a cortical region could be a a single cortical column or multiple). Our model uniquely maps regions of the input space to related regions in the columnar space based on input statistics \cite{SptialPooler, Dauletkhanuly2020}. The key insight here is that similar input regions must be mapped to adjacent regions ( or overlap) on the columnar side. A region is a soft cluster of structurally similar nodes that mimics the cortical analog of a topographic region in the sensory cortex that responds preferentially to a particular subset of inputs. Our architecture drives 300 regions \cite{Hawkins2017ATO} from the same random walks that drive the PPMI modulator. The Architecture runs Algorithm 1, which combines all three mechanisms to help WTA operate at the columnar level to determine the bias at every winner-take-all step. This algorithm can be precomputed and streamed or could be adaptive for temporal embedding with minor changes. 
\subsubsection*{\textbf{NRW} - Node to Regions Weights}
$NRW \in \mathbb{R}^{N \times R}$ assigns each node $v$ a soft probability over the $R = 300$ regions. The construction as in Equation \ref{NRW_eq} is:
\begin{equation}
\label{NRW_eq}
    NRW[v,r] = \begin{cases}
    \frac{exp \big( cos(emb[v], centroids[r]) \big)/ \tau_{NRW}}{\sum_{r' \in top-150(c)}exp \big( cos(emb[v], centroids[r'])\big)/\tau_{NRW}}   & if r \in top-150(v) \\
    0                                             & otherwise
    \end{cases}
\end{equation}
Where top-150(v) is the set of 150 regions' centroids nearest to v in cosine distance, and $\tau_{NRW} = 0.10$ is the soft assignment temperature. Each row sums to one, and each row has exactly 150 non-zero entries out of 300. Conceptually, $NRW[v,:]$ is the region node $v$ belongs to with a fuzzy community membership profile.

\subsubsection*{\textbf{CRA} - \textbf{C}olumns to \textbf{R}egions \textbf{A}ffinity}
$CRA \in \mathbb{R}^{M \times R}$ assigns each cortical mini-column $c$ a soft probability distribution over the $R = 300$ regions. The construction proceeds in two steps:
\begin{itemize}
    \item \textbf{Assign a primary region} to each column. we follow a convention that: $primary(c) = \pi(c \text{ mod } R)$ for some fixed permutation $\pi \text{ of } \{ 0,1,....R-1\}$ with $M= R =300$ which enforces every column to have a unique region. The permutation $\pi$ is an architectural choice held fixed throughout training and not learned.
    \item \textbf{Softmax-decay across regions} from the primary region,  weighted by region similarity:
    \begin{align}
        CRA[c,r] = \frac{exp \big( region_{similarity}[primary(c), r]/\tau_{CRA}\big)}{\sum_{r'}exp\big( region_{similarity}[primary(c), r']/\tau_{CRA}\big)} &  \text{  where    } \tau_{CRA} = 0.20
    \end{align}
        
\end{itemize}
Each row sums up to one. The primary region typically receives $\approx 45\%$ of the mass of the row, and the remaining mass falls across similar regions according to centroid geometry. CRA is asymmetric because different primary regions produce different row patterns in contrast to region similarity. 
\subsubsection*{\textbf{ELIG} - Eligibility and the column-selection rule}
The eligibility matrix $ELIG \in \mathbb{R}^{N \times M}$ is the matrix product:
\begin{equation}
    ELIG = NRW \cdot CRA^T.
\end{equation}
\begin{equation}
    ELIG[v,c] = \sum_{r=1}^{R} NRW[v,r] \cdot CRA[c,r]
\end{equation}
We then normalize all rows so that $max_c ELIG[v,c] = 1$ for every node $ v$. The intuition is multiplicative routing, where node $v$ belongs to a distribution over regions, and each column $c$ prefers a different distribution over regions. \textbf{Then the inner product measures how well those two distributions overlap. A high $ELIG[v,c]$ means the regions $v$ "belongs to" are exactly the regions $c$ is designed for and $c$ should compete strongly for node $v$'s activities}. A low $ELIG[v,c]$ means that the two distributions are nearly disjoint and that the column $c$ does not have interest in firing for $v$. The practical result of adopting this algorithm to bias the WTA step during training is demonstrated in Algorithm 1. The whole setup is a routing process of the current node $v$ into a region (cluster of columns) with the most competitive column wins. The association is strengthened during training steps, so that each node gets its own preferred region without being exposed to labels.

\section*{Funding Declaration}
This work was supported in part by the Center for Brain-Inspired Computing (C-BRIC), one of six centers in JUMP, a Semiconductor Research Corporation (SRC) program sponsored by the Defense Advanced Research Projects Agency (DARPA).
\section*{Code availability}
The Python implementation code for \textbf{Columnar-Embedder}architecture is hosted on \textbf{GitHub} website and available upon request. 
\section*{Data availability statement}
The datasets analyzed in this paper are publicly available standard graph benchmarks and could be accessed at \url{https://pytorch-geometric.readthedocs.io/en/2.6.0/modules/datasets.html}

\section*{Author contribution statement}
M.A. conceived the architecture, designed and implemented the model, conducted the experiments, analyzed the results, and wrote the manuscript. D.H. supervised the project, provided conceptual guidance, and reviewed and edited the manuscript. All authors reviewed and approved the final manuscript.

\appendix
\section*{Appendix}
This section introduces Algorithm 1 used to compute eligibility to bias the WTA mechanism at each cortical columnar. The computation is offline and only requires the knowledge of the PPMI matrix. The setting supports online computation because random-walk statistics can be accumulated as graph coverage increases. Thus, Algorithm 1 can be upgraded to include dynamic graph embeddings with minimal changes.   
\begin{algorithm}
\caption{Eligibility Construction of WTA for Columnar-Embedder architecture. All steps are deterministic. Given the walks and a KMeans random seed, no learning is involved.}
\label{alg:scaffold}
\begin{algorithmic}[1]
\Require Adjacency $A \in \{0,1\}^{N \times N}$; walk hyperparameters
         $(\text{wpn}, L, w)$; region count $R$; column count $M$; walks per node wpn
        walk length L;
        walk window size w;
         soft-assignment hyperparameters $(K_{\text{NRW}}, \tau_{\text{NRW}}, \tau_{\text{CRA}})$;
         column permutation $\pi$.
\Ensure Eligibility matrix $\mathrm{ELIG} \in \mathbb{R}^{N \times M}$,
        node-region weights $\mathrm{NRW} \in \mathbb{R}^{N \times R}$,
        column-region affinity $\mathrm{CRA} \in \mathbb{R}^{M \times R}$.
        
\smallskip
\Statex \textit{// Stage 1.  PPMI matrix from walk co-occurrences }
\State $\mathcal{W} \gets \mathrm{RandomWalks}(A, \text{wpn}, L)$
       \Comment{$\text{wpn}{=}10$ walks per node, length $L{=}20$}
\State Build pair stream $(u_t, v_t)_{t=1}^{T}$ from $\mathcal{W}$ with window $w{=}5$
\State $C(u) \gets \#\{t : u_t = u \text{ or } v_t = u\}$ for each $u$
\State $C(u,v) \gets \#\{t : \{u_t, v_t\} = \{u,v\}\}$ for each unordered pair
\State $\mathrm{PPMI}[u, v] \gets \max\!\left(M_{max},\, \log \dfrac{C(u, v)\,T}{C(u)\,C(v)}\right)$
       \Comment{positive PMI; sparse $N \times N$ and $M_{max} = 2$}

\smallskip
\Statex \textit{// Stage 2.  Top-64 spectral embedding }
\State $(U, \Sigma, V^{\!\top}) \gets \mathrm{TruncatedSVD}(\mathrm{PPMI},\, k{=}64)$
\State $\mathrm{emb} \gets U \cdot \sqrt{\Sigma}$
       \Comment{node spectral coordinates, $\mathrm{emb} \in \mathbb{R}^{N \times 64}$}
\State $\widetilde{\mathrm{emb}}[v] \gets \mathrm{emb}[v] \,/\, \lVert \mathrm{emb}[v] \rVert_2$
       \Comment{unit-normalized rows}

\smallskip
\Statex \textit{// Stage 3.  Region partition and similarity }
\State $\mathrm{labels}, \mathrm{centroids} \gets \mathrm{KMeans}(\widetilde{\mathrm{emb}},\, R{=}300)$
\State $\widetilde{\mathrm{centroids}}[r] \gets \mathrm{centroids}[r] \,/\, \lVert \mathrm{centroids}[r] \rVert_2$
\State $\mathrm{region\_similarity} \gets \widetilde{\mathrm{centroids}} \ \cdot \widetilde{\mathrm{centroids}}^{\!\top}$
       \Comment{Regions similarity $\in$ $\mathbb{R}^{R \times R}$, symmetric}

\smallskip
\Statex \textit{// Stage 4.  Node-region weights (NRW)}
\For{each node $v = 1, \ldots, N$}
    \State $s[v, r] \gets \cos\!\bigl(\widetilde{\mathrm{emb}}[v],\, \widetilde{\mathrm{centroids}}[r]\bigr)$
           for $r = 1, \ldots, R$
    \State $\mathcal{T}(v) \gets \mathrm{TopIndices}\bigl(s[v, :],\, K_{\text{NRW}}{=}150\bigr)$
    \For{$r = 1, \ldots, R$}
        \If{$r \in \mathcal{T}(v)$}
            \State $\mathrm{NRW}[v, r] \gets \dfrac{\exp\bigl(s[v,r] / \tau_{\text{NRW}}\bigr)}
                                       {\sum_{r' \in \mathcal{T}(v)} \exp\bigl(s[v,r'] / \tau_{\text{NRW}}\bigr)}$
                \Comment{$\tau_{\text{NRW}} = 0.10$}
        \Else
            \State $\mathrm{NRW}[v, r] \gets 0$
        \EndIf
    \EndFor
\EndFor

\smallskip
\Statex \textit{// Stage 5.  Column-region affinity (CRA)}
\State $\mathrm{primary}(c) \gets \pi(c \bmod R)$ for $c = 1, \ldots, M$
       \Comment{fixed permutation, not learned}
\For{each column $c = 1, \ldots, M$}
    \State $\mathbf{q} \gets \mathrm{region\_similarity}[\mathrm{primary}(c), :]$
    \State $\mathrm{CRA}[c, r] \gets \dfrac{\exp(\mathbf{q}[r] / \tau_{\text{CRA}})}
                                            {\sum_{r'} \exp(\mathbf{q}[r'] / \tau_{\text{CRA}})}$
           for $r = 1, \ldots, R$
           \Comment{$\tau_{\text{CRA}} = 0.20$}
\EndFor

\smallskip
\Statex \textit{// Stage 6.  Eligibility}
\State $\mathrm{ELIG} \gets \mathrm{NRW} \cdot \mathrm{CRA}^{\!\top}$
       \Comment{matrix product, $\mathbb{R}^{N \times M}$}
\State $\mathrm{ELIG}[v, :] \gets \mathrm{ELIG}[v, :] \,/\, \max_c \mathrm{ELIG}[v, c]$
       for each $v$
       \Comment{row-normalise so $\max_c \mathrm{ELIG}[v,c] = 1$}

\State \Return $\mathrm{ELIG},\, \mathrm{NRW},\, \mathrm{CRA}$
\end{algorithmic}
\end{algorithm}

\clearpage
\bibliography{references.bib}

\end{document}